\documentclass[jimaging,article,accept,moreauthors,pdftex]{Definitions/mdpi}

\usepackage{booktabs}       
\usepackage{multirow}
\usepackage{mathtools}
\usepackage{hhline}
\usepackage{chngcntr} 
\newcommand{\xdeleted}[1]{}
\newcommand{\xadded}[1]{#1}
\usepackage{xspace}
\newcommand{\STAB}[1]{\begin{tabular}{@{}c@{}}#1\end{tabular}}

\DeclarePairedDelimiterX{\infdivx}[2]{(}{)}{%
#1\;\delimsize|\delimsize|\;#2%
}
\newcommand{\kld}[2]{\ensuremath{KL\infdivx{#1}{#2}}\xspace}

\usepackage[defaultcolor=green!80!black]{changes}

\DeclarePairedDelimiter\floor{\lfloor}{\rfloor}

\usepackage{tikz}
\usepackage{collcell}

\newcommand*{\MinNumber}{0}%
\newcommand*{\MidNumber}{96} %
\newcommand*{\MaxNumber}{100}%

\newcommand{\ApplyGradient}[1]{%
\ifdim #1 pt > \MidNumber pt
\pgfmathsetmacro{\PercentColor}{max(min(100.0*(#1 - \MidNumber)/(\MaxNumber-\MidNumber),100.0),0.00)} %
\hspace{-0.33em}\colorbox{white!\PercentColor!white}{#1}
\else
\pgfmathsetmacro{\PercentColor}{max(min(100.0*(\MidNumber - #1)/(\MidNumber-\MinNumber),100.0),0.00)} %
\hspace{-0.33em}\colorbox{red!\PercentColor!white}{#1}
\fi
}

\newcolumntype{R}{>{\collectcell\ApplyGradient}c<{\endcollectcell}}
\firstpage{1}
\pubvolume{8}
\issuenum{4}
\articlenumber{93}
\pubyear{2022}
\copyrightyear{2022}
\externaleditor{Academic Editors: Simone Palazzo and Carmelo Pino
} 
\datereceived{23 January 2022}
\dateaccepted{26 March 2022}
\datepublished{31 March 2022}
\hreflink{https://doi.org/10.3390/\linebreak jimaging8040093} 

\Title{Unified Probabilistic Deep Continual Learning through Generative Replay and Open Set Recognition}

\TitleCitation{Unified Probabilistic Deep Continual Learning through Generative Replay and Open Set Recognition}

\Author{Martin Mundt $^{1,}$*$^{,\dagger}$\href{https://orcid.org/0000-0003-1639-8255}{\orcidicon}, Iuliia Pliushch $^{1}$, Sagnik Majumder $^{2}$, Yongwon Hong $^{3}$ and Visvanathan Ramesh $^{1}$\href{https://orcid.org/0000-0002-8842-905X}{\orcidicon}}

\AuthorNames{Martin Mundt, Iuliia Pliushch, Sagnik Majumder, Yongwon Hong and Visvanathan Ramesh}

\AuthorCitation{Mundt, M.; Pliushch, I.; Majumder, S.; Hong, Y.; Ramesh, V.}

\address{%
$^{1}$ \quad  Department of Computer Science and Mathematics, Goethe University, 60323 Frankfurt am Main, Germany; pliushch@em.uni-frankfurt.de (I.P.); vramesh@em.uni-frankfurt.de (V.R.) \\
$^{2}$ \quad Department of Computer Science, The University of Texas at Austin, Austin, TX 78712, USA; sagnik@cs.utexas.edu
\\
$^{3}$ \quad Department of Computer Science, Yonsei University, Seoul 03722, Korea; yhong@yonsei.ac.kr}
\firstnote{Current affiliation: Department of Computer Science, TU Darmstadt, 64289 Darmstadt, Germany.}

\corres{Correspondence: martin.mundt@tu-darmstadt.de}

\abstract{Modern deep neural networks are well known to be brittle in the face of unknown data instances and recognition of the latter remains a challenge. Although it is inevitable for continual-learning systems to encounter such unseen concepts, the corresponding literature appears to nonetheless focus primarily on alleviating catastrophic interference with learned representations. In this work, we introduce a probabilistic approach that connects these perspectives based on variational inference in a single deep autoencoder model. Specifically, we propose to bound the approximate posterior by fitting regions of high density on the basis of correctly classified data points.
These bounds are shown to serve a dual purpose: unseen unknown out-of-distribution data can be distinguished from already trained known tasks towards robust application. Simultaneously, to retain already acquired knowledge, a generative replay process can be narrowed to strictly in-distribution samples, in order to significantly alleviate catastrophic interference.}


\begin{document}


\section{Introduction}
Consider an empirically optimized deep neural network for a particular task, for the sake of simplicity, say the classification of dogs and cats. Typically, such a system is trained in a closed world setting \cite{Boult2019} according to an isolated learning paradigm \cite{Chen2016}. That is, we assume the observable world to consist of a finite set of known instances of dogs and cats, where training and evaluation is limited to the same underlying statistical data population. The training process is treated in isolation, i.e., the model parameters are inferred from the entire existing dataset at all times. However, the real world requires dealing with sequentially arriving tasks and data originating from potentially unknown sources.

In particular, should we wish to apply and extend the system to an open world, where several other animals (and non animals) exist, there are two critical questions: (a) How can we prevent obvious mispredictions if the system encounters a new class? (b) How can we continue to incorporate this new concept into our present system without full retraining? With respect to the former question, it is well known that neural networks yield overconfident mispredictions in the face of unseen unknown concepts \cite{Matan1990}, a realization that has recently resurfaced in the context of various deep neural networks \cite{Hendrycks2019, Ovadia2019, Nalisnick2019}. With respect to the latter question, it is similarly well known that neural networks, which are trained exclusively on newly arriving data, will overwrite their representations and thus forget encoded knowledge---a phenomenon referred to as catastrophic interference or catastrophic forgetting \cite{McCloskey1989, Ratcliff1990}.
Although we have worded the above questions in a way that naturally exposes their connection: to identify what is new and think about how new concepts can be incorporated, they are largely subject to separate treatment in the respective literature. While open-set recognition \cite{Scheirer2013, Scheirer2014, Boult2019} aims to explicitly identify novel inputs that deviate with respect to already observed instances, the existing continual learning literature predominantly concentrates its efforts on finding mechanisms to alleviate catastrophic interference (see \cite{Parisi2019} for an algorithmic survey).

In particular, the indispensable system component to distinguish seen from unseen unknown data, both as a guarantee for robust application and to avoid the requirement of explicit task labels for prediction, is generally missing from recent continual-learning works. Inspired by this gap, we set out to connect open-set recognition and continual learning. The underlying connecting element is motivated from the prior work of \citet{Bendale2016}, who  proposed to leverage extreme value theory (EVT) to address open-set detection in deep neural networks. The authors suggested to modify softmax prediction scores on the basis of feature space distances in blackbox discriminative models. Although this approach is promising, it alas comes with the substantial caveat that purely discriminative networks are prone to encode noise as features \cite{Ilyas2019} or fall for a most simple discriminative solution that neglects meaningful features \cite{Shah2020}. Inspired by these former insights, we set out to connect open-set recognition and continual learning, while overcoming present limitations through treatment from a generative modeling perspective.

Our specific contributions are that we propose to unify the prevention of catastrophic interference in continual learning with open-set recognition in a single model. Specifically, we extend prior EVT works \cite{Scheirer2013, Scheirer2014, Bendale2016} to a natural formulation on the basis of the aggregate posterior in variational inference with deep autoencoders \cite{Kingma2013, Kingma2015}. By \textit{identifying out-of-distribution instances} we can detect unseen unknown data and prevent false predictions; by explicitly \textit{generating in-distribution samples} from areas of high probability density under the aggregate posterior, we can simultaneously circumvent rehearsal of ambiguous uninformative examples. This leads to robust application, while significantly reducing catastrophic interference. We empirically corroborated our approach in terms of improved out-of-distribution detection performance and simultaneously reduced the continual catastrophic interference.  We further demonstrate benefits through recent deep generative modeling advances, such as autoregression \cite{Oord2016, Gulrajani2017, Chen2016} and introspection \cite{Huang2018, Ulyanov2018}, validated by scaling to high-resolution color images.

\subsection{Background and Related Work}
\subsubsection{Continual Learning}
In isolated supervised learning, the core assumption is the presence of i.i.d.~data at all times and training is conducted using a dataset $\boldsymbol{D} \equiv \left\{ \left( \boldsymbol{x}^{(n)}, y^{(n)} \right) \right\}_{n=1}^{N}$, consisting of $N$ pairs of data instances $\boldsymbol{x}^{(n)}$ and their corresponding labels $y^{(n)} \in \left\{ 1 \ldots C \right\}$ for $C$ classes. In contrast, in continual learning, data $\boldsymbol{D}_{t} \equiv \left\{ \left( \boldsymbol{x}_{t}^{(n)}, y_{t}^{(n)} \right) \right\}_{n=1}^{N_{t}}$ with $t = 1, \ldots, T$ arrives sequentially for $T$ disjoint sets, each with number of classes $C_{t}$.

It is assumed that only the data of the current task is available. Without additional mechanisms, tuning on such a sequence will lead to catastrophic interference \cite{McCloskey1989, Ratcliff1990}, i.e.,~representations of former tasks being overwritten through present optimization. A recent review of many continual-learning algorithms to prevent said interference was provided by Parisi et al. \cite{Parisi2019}. Here, we present a brief summary of the key underlying principles.

Alleviating catastrophic interference is most prominently addressed from two angles. Regularization methods, such as synaptic intelligence (SI) \cite{Zenke2017} or elastic weight consolidation (EWC) \cite{Kirkpatrick2017} explicitly constrain the weights during continual learning to avoid drifting too far away from the previous tasks' solutions. In a related picture, learning without forgetting~\cite{Li2016} uses knowledge distillation \cite{Hinton2014} to regularize the end-to-end functional.

Rehearsal methods on the other hand, store data subsets from distributions belonging to old tasks or generate samples in pseudo-rehearsal \cite{Robins1995}. The central component of the latter is thus the selection of significant instances. For methods, such as incremental classifier and representation learning (iCarl) \cite{Rebuffi2017}, it is therefore common to resort to auxiliary techniques, such as  the nearest-mean classifier \cite{Mensink2012} or core sets~\cite{Bachem2015}. Inspired by complementary learning systems \cite{OReilly2003}, dual-model approaches sample data from a separate generative memory. In a bio-inspired incremental learning architecture (GeppNet) \cite{Gepperth2016}, long short-term memory \cite{Hochreiter1997} is used for storage, whereas generative replay \cite{Shin2017} samples from an additional generative adversarial network (GAN) \cite{Goodfellow2014}.

As detailed in Variational Generative Replay (VGR) \cite{Farquhar2018, Farquhar2018a}, methods with a Bayesian perspective encompass a natural capability for continual learning by making use of the learned distribution. Existing works nevertheless fall into the above two categories and their combination: a prior-based approach using the former task's approximate posterior as the new task's prior \cite{Nguyen2018} or estimating the likelihood of former data through generative replay or other forms of rehearsal \cite{Farquhar2018, Achille2018}. Crucially, the success of many continual-learning techniques can be attributed primarily to the considered evaluation scenario. With the exception of VGR \cite{Farquhar2018}, the majority of above techniques train a separate classifier per task and thus either require the explicit storage of task labels or assume the presence of a task oracle during evaluation. This multi-head scenario prevents ``cross-talk'' between classifier units by not sharing them, which would otherwise rapidly decay the accuracy as newly introduced classes directly confuse existing concepts. While the latter is acceptable to limit catastrophic interference, it also signifies a major limitation in practical applications. Even though VGR \cite{Farquhar2018} uses a single classifier, the researchers trained a separate generative model per task to avoid catastrophic interference in the generator.

Our approach builds upon these previous works and leverages variational inference in deep generative models. However, we propose to tie the prevention of catastrophic interference with open-set recognition through a natural mechanism based on the aggregate posterior in a single model.

\subsubsection{Out-of-Distribution and Open Set Recognition}
The above-mentioned literature focused their efforts predominantly on addressing catastrophic interference. Even though continual learning is the desideratum, the corresponding evaluation is thus conducted in a closed world setting, where instances that do not belong to the observed data distribution are not encountered. In reality, this is not guaranteed as users could provide arbitrary inputs or unknowingly present the system with novel inputs that deviate substantially from previously seen instances. Our models thus require the ability to identify unseen examples in the unconstrained open world and categorize them as either belonging to the already known set of classes or as presently being unknown. We provide a small overview of approaches that aim to address this question in deep neural networks. A comprehensive survey was provided by Boult et al. \cite{Boult2019}.

As the most simple approach, the aim of calibration works is to separate a known and unknown input through prediction confidence, often by fine tuning or re-training an already existing model. In out-of-distribution detector for neural networks (ODIN) \cite{Liang2018}, this is addressed through perturbations and temperature scaling, while Lee et al. \cite{Lee2018a} used a separately trained GAN to generate out-of-distribution samples from low probability densities and explicitly reduced their confidence through the inclusion of an additional loss term. Similarly, the objectosphere loss \cite{Dhamija2018} defines an objective that explicitly aims to maximize entropy for upfront available unknown inputs.

As we do not have access to future data a priori, by definition, a naive conditioning or calibration on unseen unknown data is infeasible.  The commonly applied thresholding is insufficient as overconfident prediction values cannot be prevented \cite{Matan1990}. Bayesian neural network models \cite{MacKay1992} could be believed to intrinsically be able to reject statistical outliers through model uncertainty \cite{Farquhar2018} and overcome this limitation of overconfident prediction values. For use with deep neural networks, it was suggested that stochastic forward passes with Monte-Carlo Dropout (MCD) \cite{Gal2015} can provide a suitable approximation. However, the closed-world assumption in training and evaluation still persists \cite{Boult2019}. In addition, variational approximations in deep networks \cite{Graves2011, Kingma2013, Farquhar2018, Achille2018} and corresponding uncertainty estimates suffer from similar overconfidence, and the  distinction of unseen out-of-distribution data from already trained knowledge is known to be unsatisfactory \cite{Nalisnick2019, Ovadia2019}.

A more formal approach was suggested in works based on open-set recognition~\cite{Scheirer2013}. The key here is to limit predictions originating from open space, that is, the area in obtained embeddings that is outside of a small radius around previously observed training examples. Without re-training, post hoc calibration or modifying loss functions, one approach to open-set recognition in deep networks is through extreme-value theory (EVT) \cite{Scheirer2014, Bendale2016}. Here, limiting the threat of overconfidence is based on monotonically decreasing the recognition function's probability with respect to increasing distance of instances to the feature embedding of known training points. The Weibull distribution, as one member of the family of extreme value distributions, has been empirically demonstrated to work well in conjunction with distances in the penultimate deep network layer as the underlying feature space. On the basis of extreme values to this layer's average activation values, the authors devised a procedure to revise the Softmax prediction values, referred to as~OpenMax.

In a similar spirit, our work avoids relying on predictive values, while also moving away from empirically chosen deep neural network feature spaces. We instead propose to use EVT to bind the approximate posterior in variational inference. We thus directly operate on the underlying (lower-bound to the) data distribution and the generative factors. This additionally allows us to constrain the generative replay to distribution inliers, which further alleviates catastrophic interference.

\section{Materials and Methods}

\subsection{Unifying Catastrophic Interference Prevention with Open Set Recognition}
We first summarize the preliminaries on continual learning from a perspective of variational inference in deep generative models \cite{Graves2011, Kingma2013}. We then proceed by bridging the improved prevention of catastrophic interference in continual learning with the detection of unseen unknown data in open-set recognition.

\subsubsection{Preliminaries: Learning Continually through Variational Auto-Encoding}
\xadded{We start with a problem scenario similar to the one introduced in ``Auto-Encoding Variational Bayes'' \cite{Kingma2013}, i.e., we assume that there exists a data generation process responsible for the creation of the labeled data given some random latent variable $\boldsymbol{z}$. We} consider a model with a shared encoder with variational parameters $\boldsymbol{\theta}$, decoder and \textit{linear} classifier with respective parameters $\boldsymbol{\phi}$ and $\boldsymbol{\xi}$. The joint probabilistic encoder learns an encoding to a latent variable $\boldsymbol{z}$, over which a unit Gaussian prior $p(\boldsymbol{z})$ is placed.

\begin{figure}[H]

\includegraphics[width = 0.875\columnwidth]{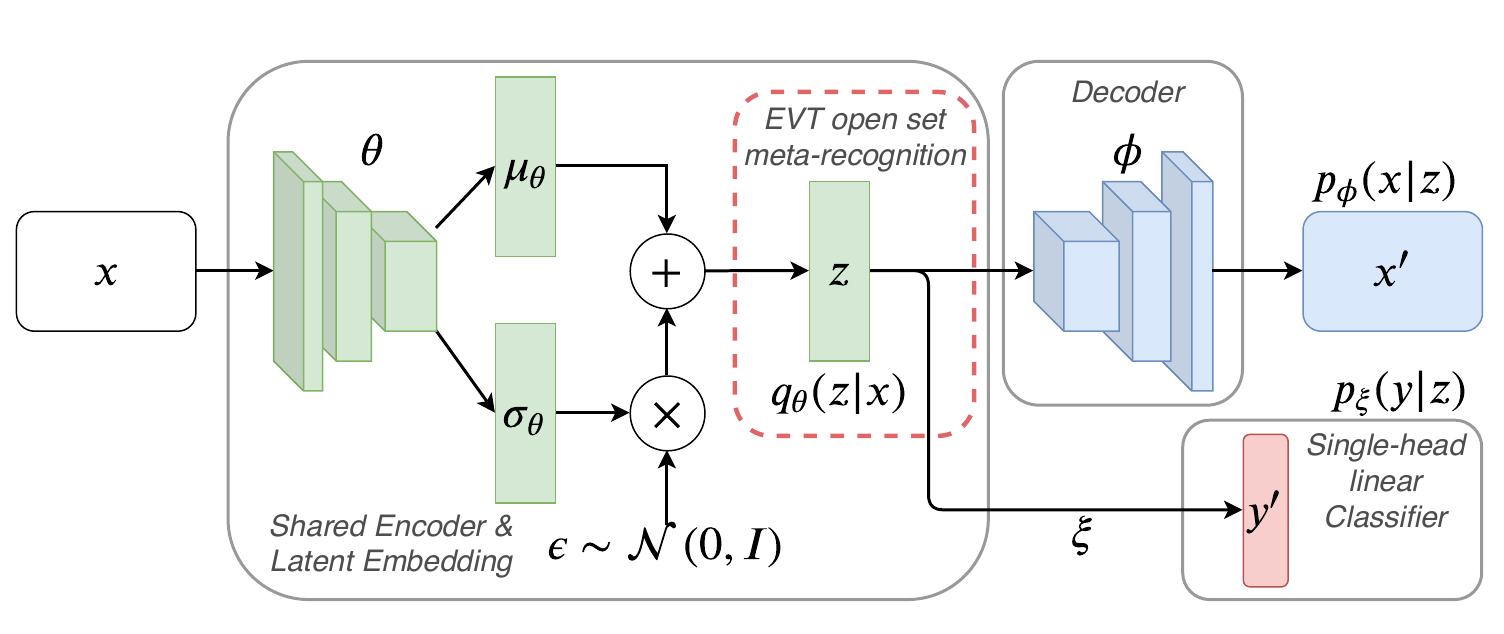}
\caption{A joint continual-learning model consisting of a shared probabilistic encoder $q_{\boldsymbol{\theta}}(\boldsymbol{z}|\boldsymbol{x})$, probabilistic decoder $p_{\boldsymbol{\phi}}(\boldsymbol{x}, \boldsymbol{z})$ and probabilistic classifier $p_{\boldsymbol{\xi}}(y, \boldsymbol{z})$. For open-set recognition and generative replay with outlier rejection, extreme-value theory (EVT) based bounds on the basis of the approximate posterior \mbox{are~established.}}\label{fig:CLArchitecture}
\end{figure}

Using variational inference, the encoder's purpose is to approximate the true posterior to $p_{\boldsymbol{\phi}}(\boldsymbol{x}, \boldsymbol{z})$ and $p_{\boldsymbol{\xi}}(y, \boldsymbol{z})$. The probabilistic decoder $p_{\boldsymbol{\phi}}(\boldsymbol{x} | \boldsymbol{z})$ and probabilistic linear classifier $p_{\boldsymbol{\xi}}(y | \boldsymbol{z})$ then return the conditional probability density of the input $\boldsymbol{x}$ and target $y$ under the respective generative model given a sample $\boldsymbol{z}$ from the approximate posterior $q_{\boldsymbol{\theta}}(\boldsymbol{z}|\boldsymbol{x})$. This yields a generative model $p(\boldsymbol{x}, \boldsymbol{y}, \boldsymbol{z})$, for which we assume a factorization and generative process of the form $p(\boldsymbol{x}, \boldsymbol{y}, \boldsymbol{z}) = p(\boldsymbol{x}|\boldsymbol{z})p(\boldsymbol{y}|\boldsymbol{z})p(\boldsymbol{z})$. For variational inference with this model, the sum over all elements in the dataset n $\in$ D in the following lower-bound is optimized:
\begin{equation} \label{eq:general_loss}
p(\boldsymbol{x}, {\boldsymbol{y}}) \geq  \mathbb{E}_{q_{\boldsymbol{\theta}}(\boldsymbol{z} | \boldsymbol{x})} \left[ \log{p_{\boldsymbol{\phi}}(\boldsymbol{x} | \boldsymbol{z})}  {+ \log{p_{\boldsymbol{\xi}}(\boldsymbol{y} | \boldsymbol{z})}} \right] - {\beta} \kld{q_{\boldsymbol{\theta}}(\boldsymbol{z} | \boldsymbol{x})}{p(\boldsymbol{z})} \, ,
\end{equation}
where \emph{KL} denotes the Kullback-Leibler divergence. In other words, the right hand side of Equation \eqref{eq:general_loss} defines our loss $\mathcal{L}\left(\boldsymbol{x}, \boldsymbol{y}; \boldsymbol{\theta}, \boldsymbol{\phi}, \boldsymbol{\xi} \right)$. This model can be seen as employing a variant of a (semi-)supervised variational auto-encoder (VAE) \cite{Kingma2015} with a $\beta$ term \cite{Higgins2017}, where, in addition to approximating the data distribution, the model learns to incorporate the class structure into the latent space. Without the blue terms, the original unsupervised VAE formulation \cite{Kingma2013} is recovered. This forms the basis for continual learning with open-set recognition as discussed in the subsequent section.
An illustration of the model is shown in Figure \ref{fig:CLArchitecture}.

\xadded{Abstracting away from the mathematical detail and speaking informally about the intuition behind the model, we first encode a data input $\boldsymbol{x}$ and encode it into two vectors. These vectors represent the mean and standard deviation of a Gaussian distribution. Using the reparametrization trick $\varepsilon \cdot \sigma + \mu$, a sample from this distribution is then calculated. During training, the respective embedding, also referred to as the latent space, is encouraged to follow a unit Gaussian distribution through the minimization of the Kullback-Leibler divergence. A linear classifier that operates directly on this latent embedding to predict a class for a sample additionally ensures that the obtained distribution is clustered according to the classes.

Examples of such fits are shown in the later Figure \ref{fig:Latent_2D}. Finally, the decoder takes, as input, the latent variable and reconstructs the original data input during training. Once the model is finished training, we can also directly draw a sample from the Gaussian distribution, obtain a latent sample and generate a novel data point directly, without the need to compute the encoder first.} \xdeleted{and the} \xadded{A} corresponding full \xadded{and formal} derivation of Equation~\eqref{eq:general_loss}, the lower-bound to the joint distribution $p(\boldsymbol{x}, \boldsymbol{y})$ is supplied in Appendix \ref{App_sec_lowerbound}.

\vspace{-12pt}
Without further constraints, one could continually train the above model by sequentially accumulating and optimizing Equation \eqref{eq:general_loss} over all currently present tasks $t = 1, \ldots, T$.
Being based on the accumulation of real data, this provides an upper bound to the achievable performance in continual learning. However, this form of continued training is generally infeasible if only the most recent task's data is assumed to be available. Making use of the model's generative nature, we can follow previous works \cite{Farquhar2018, Achille2018} and estimate the likelihood of former data through generative replay:\vspace{-6pt}
\begin{equation}
\label{eq:CL_loss}
\mathcal{L}_{t}\left(\boldsymbol{x}, \boldsymbol{y}; \boldsymbol{\theta}, \boldsymbol{\phi}, \boldsymbol{\xi} \right) =  \frac{1}{2}
\frac{1}{N_{t}}  \sum_{n=1}^{N_{t}} \mathcal{L}\left(\boldsymbol{x}_{t}^{(n)}, \boldsymbol{y}_{t}^{(n)}; \boldsymbol{\theta}, \boldsymbol{\phi}, \boldsymbol{\xi} \right) + \frac{1}{2} \frac{1}{N^{\prime}_{t}}  \sum_{n=1}^{N_{t}^{\prime}} \mathcal{L}\left(\boldsymbol{{x}}_{t}^{\prime (n)}, \boldsymbol{y}_{t}^{\prime (n)}; \boldsymbol{\theta}, \boldsymbol{\phi}, \boldsymbol{\xi} \right)
\end{equation}
where
\begin{equation}\label{eq:gen_sampling}
\boldsymbol{x}_{t}^{\prime } \sim p_{\boldsymbol{\phi}, t-1}(\boldsymbol{x} | \boldsymbol{z}); \, y_{t}^{\prime} \sim p_{\boldsymbol{\xi}, t-1}(y | \boldsymbol{z}) \, \, \mathtt{and} \, \, \boldsymbol{z} \sim p(\boldsymbol{z}) \, .
\end{equation}

Here, $\boldsymbol{x}_{t}^{\prime}$ is a sample from the generative model with its corresponding classifier label $y_{t}^{\prime }$. $N_{t}^{\prime}$ is the number of instances of all previously seen tasks. In this way, the expectation of the log-likelihood for all previously seen tasks is estimated and the dataset at any point in time $\boldsymbol{\tilde{D}}_{t} \equiv \left\{ ( \boldsymbol{\tilde{x}}_{t}^{(n)}, \tilde{y}_{t}^{(n)} ) \right\}_{n=1}^{\tilde{N}_{t}} = \left\{ ( \boldsymbol{x}_{t} \cup \boldsymbol{x}_{t}^{\prime}, y_{t} \cup y_{t}^{\prime }) \right\}$ is a concatenation of past data generations and the current task's real data.

\subsubsection{Open Set Recognition and Generative Replay with Statistical Outlier Rejection}
Trained naively in the above fashion, our model will unfortunately suffer from accumulated errors with each successive iteration of generative replay, similar to the current literature approaches. To avoid this, we would alternatively require the training of multiple encoders to approximate each task's posterior individually, as in variational continual learning (VCL) \cite{Nguyen2018}, or train multiple generators, as in VGR \cite{Farquhar2018}. We posit that the main challenge is how high-density areas under the prior $p(\boldsymbol{z})$ are not necessarily reflected in the structure of the aggregate posterior $q_{\boldsymbol{\theta}, t}(\boldsymbol{z})$ \cite{Tomczak2018}. The latter refers to the practically obtained encoding \cite{Hoffman2016}:\vspace{-6pt}
\begin{equation}\label{eq:aggregate_posterior}
q_{\boldsymbol{\theta}, t}(\boldsymbol{z}) = \mathbb{E}_{p_{\tilde{D}_{t}}(\boldsymbol{\tilde{x}})} \left[ q_{\boldsymbol{\theta}, t}(\boldsymbol{z} | \boldsymbol{\tilde{x}})\right] \approx \frac{1}{\tilde{N}_{t}} \sum_{n=1}^{\tilde{N}_{t}} q_{\boldsymbol{\theta}, t}(\boldsymbol{z} | \boldsymbol{\tilde{x}}^{(n)})
\end{equation}

\vspace{-8pt}
\begin{figure}[H]

\includegraphics[width = 0.3 \columnwidth]{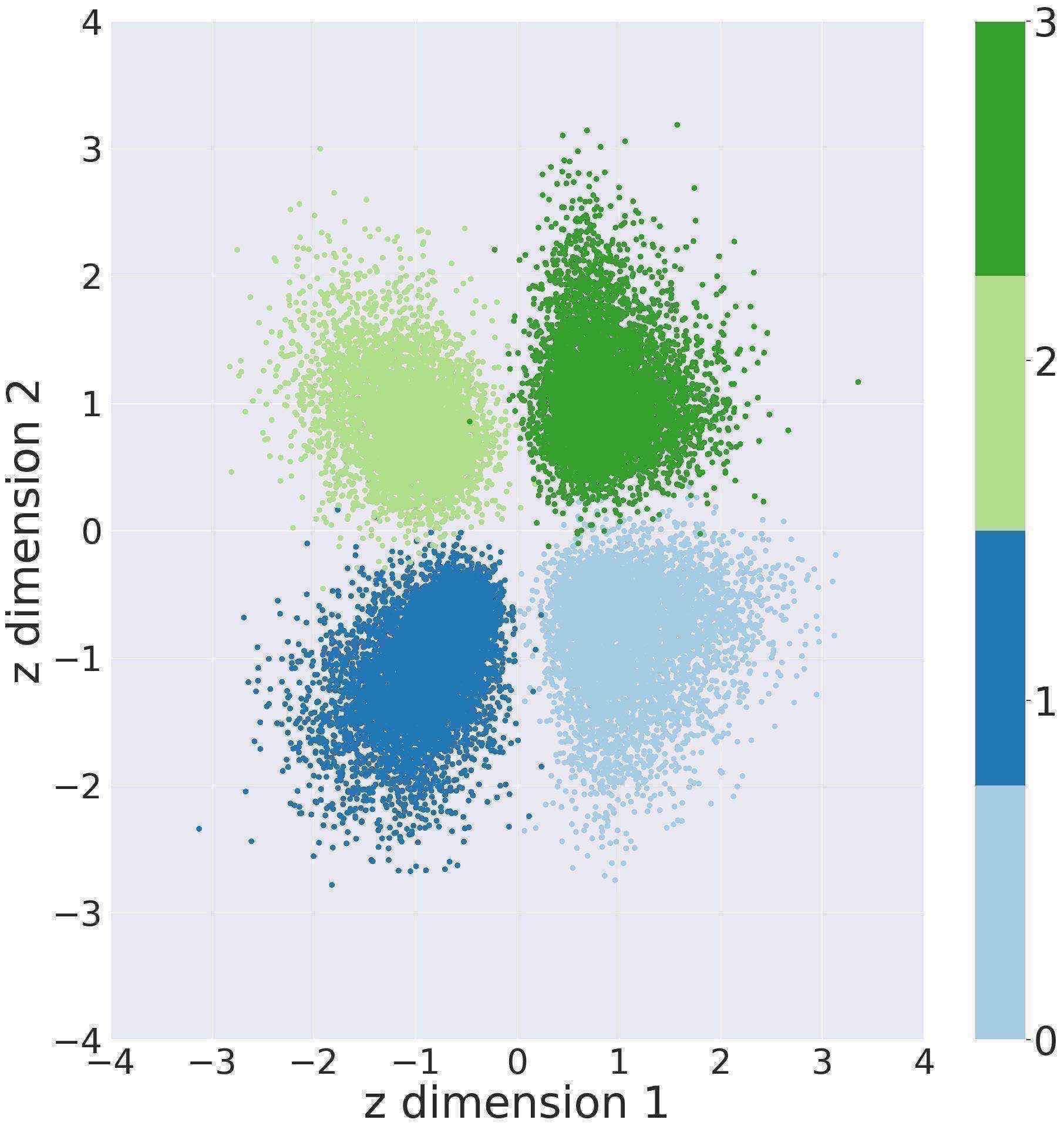}
\quad
\includegraphics[width = 0.3 \columnwidth]{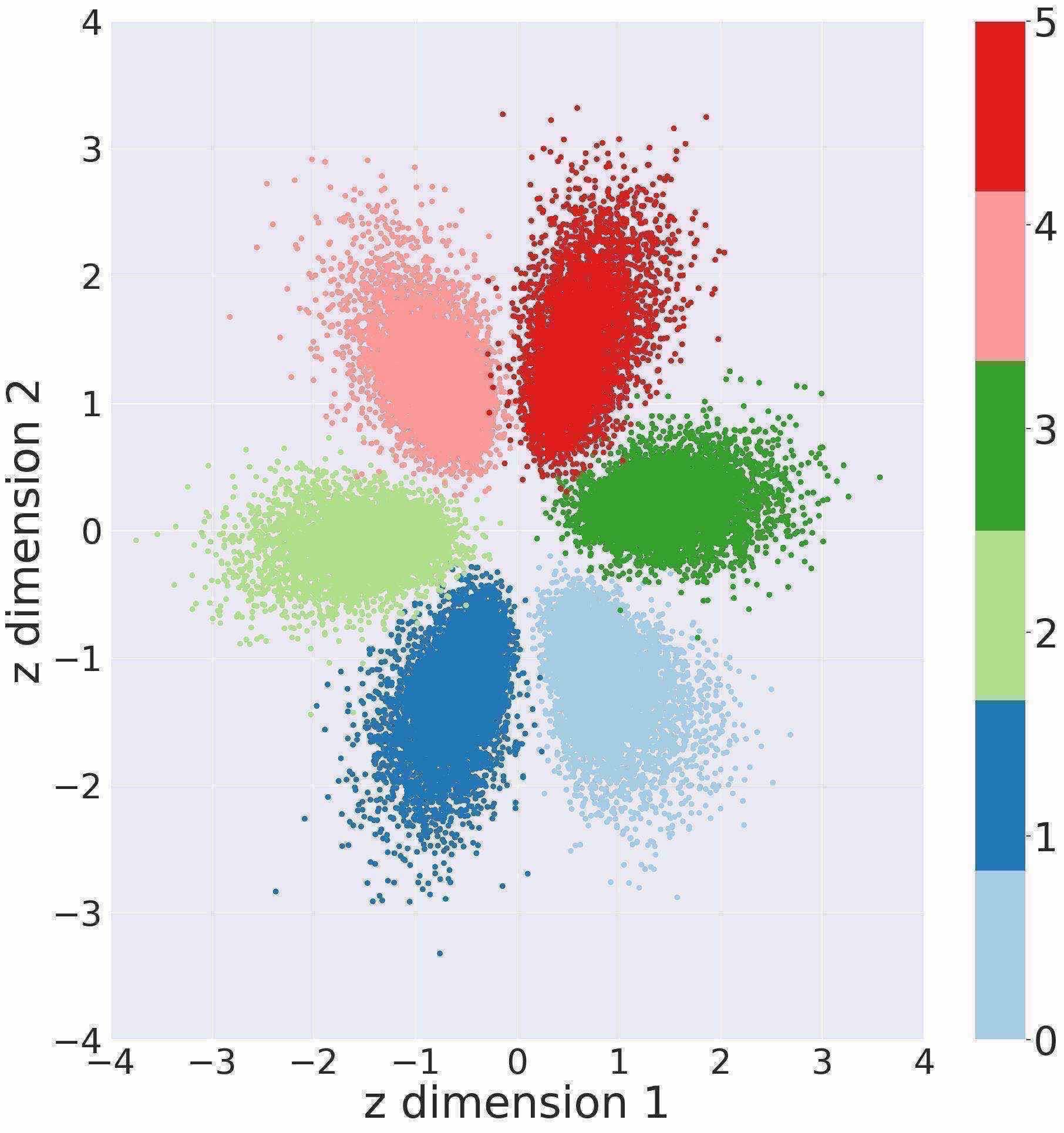}
\quad
\includegraphics[width = 0.3 \columnwidth]{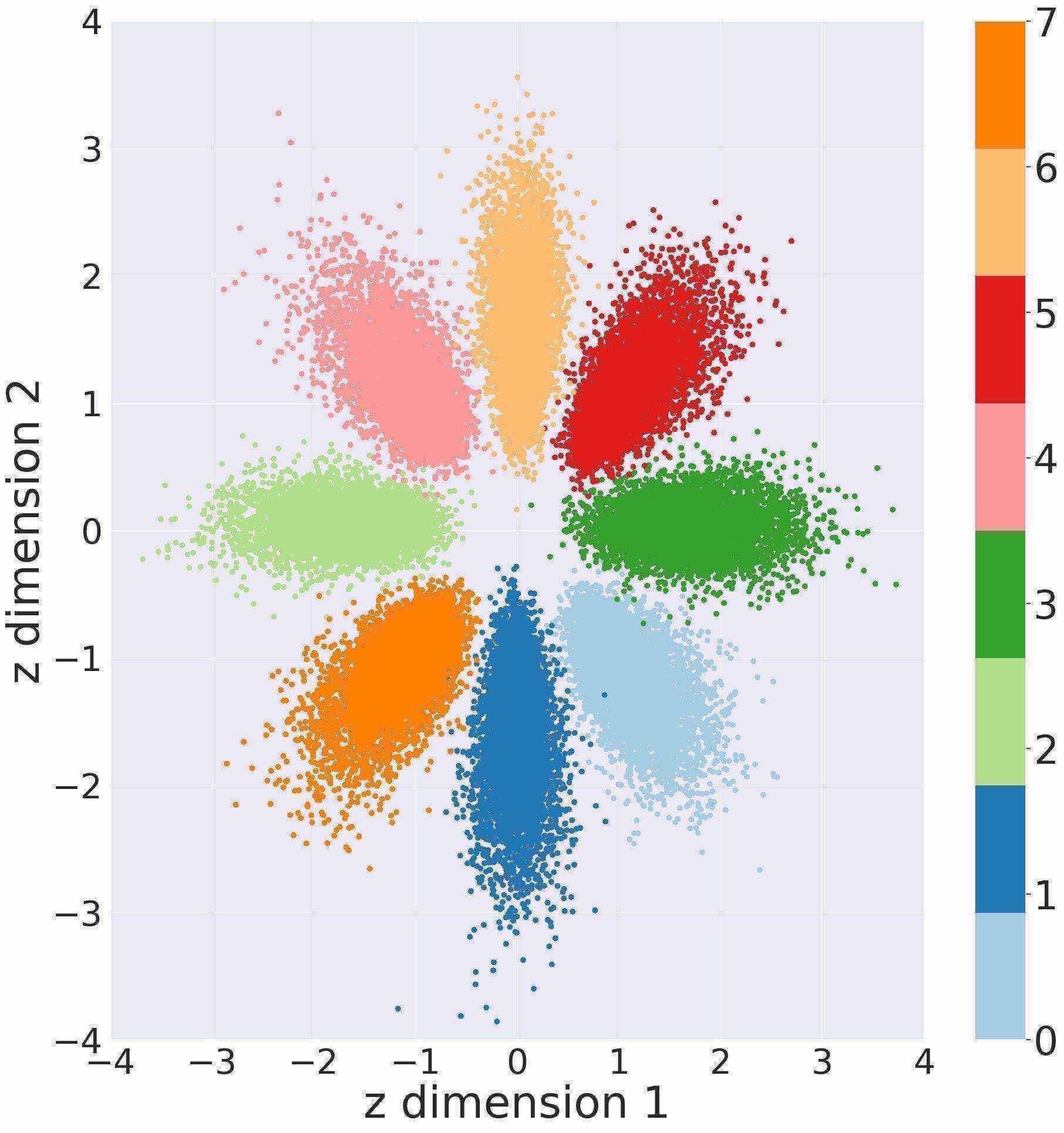}
\caption{\label{fig:Latent_2D} 2-D latent space aggregate posterior visualization for continually learned MNIST (Modified National Institute of Standards and Technology database). From left to right, the latent space for four, six and then eight classes are shown. This is best viewed in color. }
\end{figure}
To provide intuition, we illustrate this prior-posterior discrepancy on the obtained two-dimensional latent encodings for a continually trained supervised MNIST (Modified National Institute of Standards and Technology database) \cite{LeCun1998} model in Figure \ref{fig:Latent_2D}. Here, we can make two observations: to preserve the inherent data structure, the aggregate posterior deviates from the prior. In fact, this is further amplified by the imposed necessity for linear class separation and the beta term in Equation \eqref{eq:general_loss}; however, we note that the discrepancy is desired even in completely unsupervised scenarios \cite{Hoffman2016, Tomczak2018}.

The underlying rationale is that there needs to be a balance in the effective latent encoding overlap \cite{Burgess2017}, which can best be summarized with a direct quote from the recent work of \mbox{Mathieu et al. \cite{Mathieu2019}:} \emph{``The overlap is perhaps best understood by considering extremes: with too little the latents effectively become a lookup table; too much, and the data and latents do not convey information about each other. In either case, meaningfulness of the latent encodings is lost." (p. 4)}. Additional discussion on the role of beta can be found in Appendix \ref{sec:appx_beta}.

Thus, the generated data from low-density regions of the aggregate posterior do not generally correspond to the encountered data instances. Conversely, data instances that fall into high-density regions under the prior should not generally be considered as statistical inliers with respect to the observed data distribution; recall Figure \ref{fig:Latent_2D}. This boundary between low- and high-density regions forms the basis for a natural connection between open-set recognition and continual learning: generate from high-density regions and reject novel instances that fall into low-density regions.

Ideally, we could find a solution by replacing the prior in the \emph{KL} divergence of \mbox{Equation \eqref{eq:general_loss}} with $q_{\boldsymbol{\theta}, t}(\boldsymbol{z})$ and, respectively, sampling $z \sim q_{\boldsymbol{\theta}, t-1}(\boldsymbol{z})$ in \mbox{Equations \eqref{eq:CL_loss} and \eqref{eq:gen_sampling}}. Even though using the aggregate posterior as a subsequent prior is the objective in multiple recent works, it can be challenging in high dimensions, lead to over-fitting or come at the expense of additional hyper-parameters \cite{Tomczak2018, Bauer2019, Takahashi2019}. To avoid finding an explicit representation for the multi-modal $q_{\boldsymbol{\theta}, t}(\boldsymbol{z})$, we draw inspiration from the EVT-based OpenMax approach \cite{Bendale2016} in deep neural networks. However, instead of using knowledge about extreme distances in penultimate layer activations to modify a Softmax prediction, we now propose to apply EVT on the basis of the class conditional aggregate posterior.

In this view, any sample can be regarded as statistically outlying if its distance to the classes' latent mean is extreme with respect to what has been observed for the majority of correctly predicted data instances, i.e., the sample falls into a region of low density under the aggregate posterior and is less likely to belong to $p_{\tilde{D}}(\boldsymbol{\tilde{x}})$.
For convenience, let us introduce the indices of all correctly classified instances at the end of task $t$ as $m = 1, \ldots, \tilde{M}_{t}$. To obtain bounds on the aggregate posterior, we first define the mean latent vector for each class for all correctly predicted seen data instances $\boldsymbol{\bar{z}}_{c, t}$ and the respective set of latent distances as
\begin{equation}\label{eq:latent_distance}
\Delta_{c, t} \equiv \left\{ f_{d} \left( \boldsymbol{\bar{z}}_{c, t}, \mathbb{E}_{q_{\theta, t}(\boldsymbol{z}|\boldsymbol{\tilde{x}}_{t}^{(m)})} \left[ \boldsymbol{z} \right] \right) \right\}_{m \in \tilde{M}_{c, t}} \quad \mathtt{with} \quad \boldsymbol{\bar{z}}_{c, t} = \frac{1}{|\tilde{M}_{c, t}|}  \sum_{m \in \tilde{M}_{c, t}} \mathbb{E}_{q_{\theta, t}(\boldsymbol{z}|\boldsymbol{\tilde{x}}_{t}^{(m)})} \left[ \boldsymbol{z} \right] \, .
\end{equation}

Here, $f_{d}$ signifies a choice of distance metric. We proceed to model this set of distances with a per class heavy-tail Weibull distribution $\boldsymbol{\rho}_{c, t} = (\tau_{c, t}, \kappa_{c, t}, \lambda_{c, t})$ on $\Delta_{c, t}$ for a given tail-size $\eta$. As these distances are based on the class conditional approximate posterior, we can thus bound the latent space regions of high density. The tightness of the bound is characterized through $\eta$, that can be seen as a prior belief with respect to the outlier quantity assumed to be inherently present in the data distribution. The choice of $f_{d}$ determines the nature and dimensionality of the obtained distance distribution. For our experiments, we find that the cosine distance and thus a univariate Weibull distance distribution per class seems to be sufficient. Using the cumulative distribution function of this Weibull model $\boldsymbol{\rho}_{t}$ we can now estimate any sample's outlier (or inlier) probability:
\begin{equation}\label{eq:outlier_probability}
\omega_{\boldsymbol{\rho}, t}(\boldsymbol{z}) =  \mathtt{min} \left( 1 - \exp \left(-  \frac{| f_{d} \left( \bar{\boldsymbol{z}}_{t}, \boldsymbol{z} \right)- \boldsymbol{\tau}_{t} |}{\boldsymbol{\lambda}_{t}} \right)^{\boldsymbol{\kappa}_{t}} \right) \quad ,
\end{equation}
where the minimum returns the smallest outlier probability across all classes. If this outlier probability is larger than a prior rejection probability $\Omega_{t}$, the instance can be considered as unknown. Such a formulation, which we term open variational auto-encoder (OpenVAE), now provides us with the means to learn continually and identify unknown data:

\begin{enumerate}
\item For a novel data instance, Equation \eqref{eq:outlier_probability} yields the outlier probability based on the probabilistic encoder $\boldsymbol{z} \sim q_{\boldsymbol{\theta, t}}(\boldsymbol{z} | \boldsymbol{x})$, and a false overconfident classifier prediction can be avoided.
\item To mitigate catastrophic interference, Equation \eqref{eq:outlier_probability} can be used on top of $\boldsymbol{z} \sim p(\boldsymbol{z})$ to constrain the generative replay (Equation \eqref{eq:gen_sampling}) to the aggregate posterior thus avoiding the need to sample it directly.
\end{enumerate}

\begin{figure}[H]

\includegraphics[width =  0.95 \columnwidth]{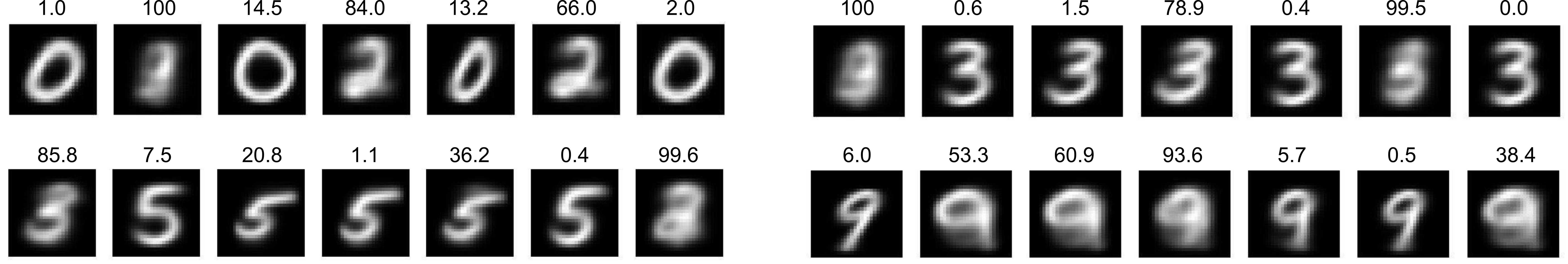} \\
\medskip
\includegraphics[width = 0.95 \columnwidth]{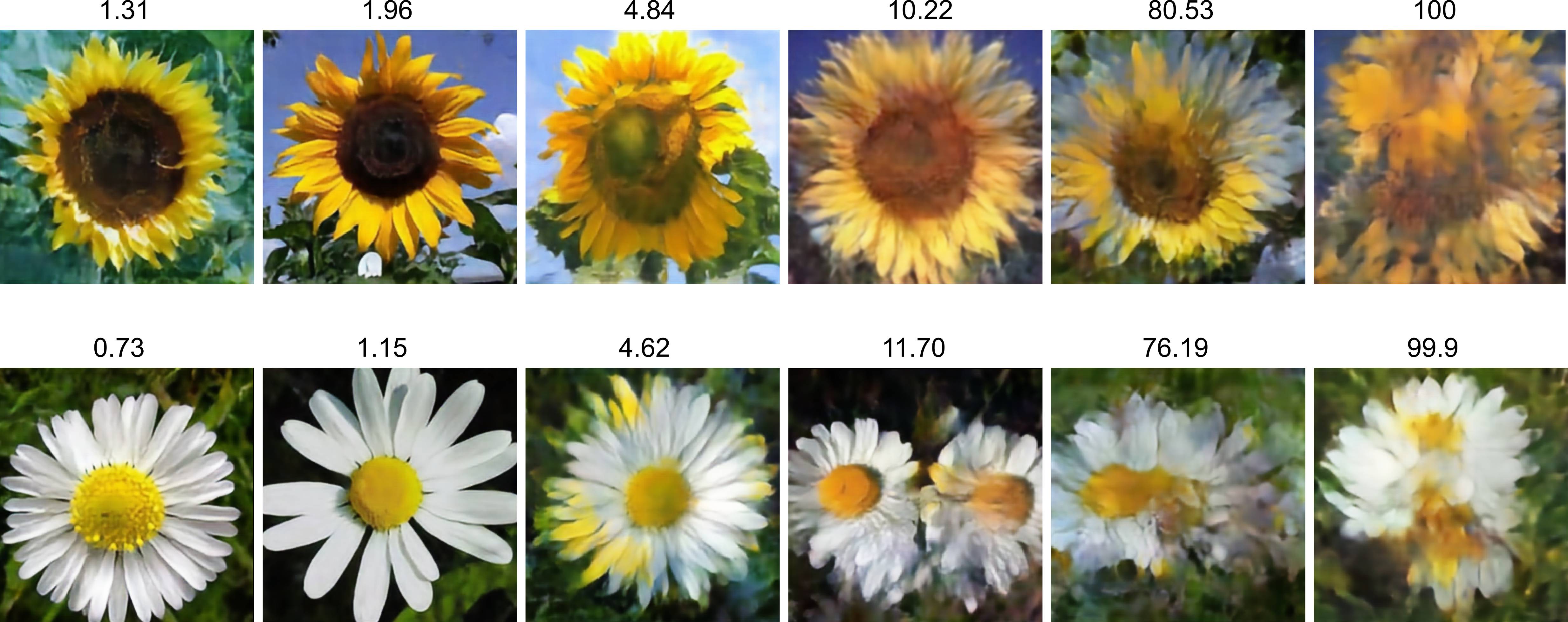}
\caption{Generated images $x \sim p_{\boldsymbol{\phi}, t}(\boldsymbol{x}|\boldsymbol{z})$ with $\boldsymbol{z} \sim p(\boldsymbol{z})$ and their corresponding class $c$ obtained from the classifier $p_{\boldsymbol{\xi}, t}(y|\boldsymbol{z})$ together with their open-set outlier percentage in our proposed open variational auto-encoder (OpenVAE). Image quality degradation and class ambiguity can be observed with the increasing outlier likelihood.  Generated $28 \times 28$ MNIST images are from the 2-D latent space of Figure \ref{fig:Latent_2D},  classified as $c = 0$ (\textbf{top left}), $c = 3$ (\textbf{top right}), $c = 5$ (\textbf{bottom left}) and $c = 9$ (\textbf{bottom right}).  Generated $256 \times 256$ resolution flower images are based on a 60-dimensional latent space of a model trained with introspection (see experiments and Appendix \ref{app:sec_gen_advances}),   which are classified as ``sunflower'' (\textbf{top}) and ``daisy'' (\textbf{bottom}).}
\label{fig:generated_openset}
\end{figure}

To give an illustration of the benefits, we show the generated MNIST \cite{LeCun1998} and larger resolution flower images \cite{Nilsback2006} together with their outlier percentage in Figure \ref{fig:generated_openset}. In practical application, we  discard the ambiguous examples that are due to low-density regions and thus a high outlier probability. Even though we conduct sampling with rejection, note how this is computationally efficient, as we only need to calculate the heavy probabilistic decoder for accepted statistically inlying examples, and sampling from the prior with computation of Equation \eqref{eq:outlier_probability} is almost negligible in comparison.

\vspace{-9pt}

\section{Results}
Instead of presenting a single experiment for continual learning in the constant presence of outlying non-task data, we chose to empirically corroborate our proposed approach in two experimental parts. The first section is dedicated to out-of-distribution detection, where we demonstrate the advantages of EVT in our generative model formulation. We then proceed to showcase how catastrophic interference is also mitigated by confining generative replay to aggregate posterior inliers in class incremental learning.

We emphasize that whereas the sections are \textit{presented individually}, our approach's uniqueness lies in \textit{using a core underlying mechanism to unify both challenges simultaneously}. The rationale behind choosing this form of presentation is to help readers better contextualize the contribution of OpenVAE with the existing literature as, to the best of our knowledge, there exists no present other work that yields adequate continual classification accuracy while being able to robustly recognize unknown data instances.  As such, we will now see that existing continual-learning approaches provide no suitable mechanism to overcome the challenge of providing robust predictions when data outside the known benchmark set are included.

\subsection{Open Set Recognition}
We experimentally highlight OpenVAE's ability to distinguish unknown task data from data belonging to known tasks to avoid overconfident false predictions.


\subsubsection*{Experimental Set-Up and Evaluation}
In summary, our goal is \xadded{two-fold. The typical goal is to train on an initial task and correctly classify the held-out or unseen test data for this task. That is, we desire a large average classification test accuracy. In addition to this, in order to ensure that this classification is robust to unknown data, we now additionally desire to have a large value for a second kind of accuracy. Our simultaneous goal is} to consider all test data of already trained tasks as inlying, while successfully identifying $100$\% of completely unknown datasets as outliers. 

For this purpose, we evaluate OpenVAE's and other models' capability to distinguish the in-distribution test set of a respectively trained MNIST (Modified National Institute of Standards and Technology database)~\cite{LeCun1998}, FashionMNIST \cite{Xiao2017}, AudioMNIST \cite{Becker2018} from the other two and several unknown datasets: Kuzushiji-MNIST (KMNIST)~\cite{Clanuwat2018}, Street-View House Numbers (SVHN)~\cite{Netzer2011} and Canadian Institute for Advanced Research (CIFAR) datasets (in both versions with 10 and 100 classes) \cite{Krizhevsky2009}. Here, the (Fourier-transformed) audio data is included to highlight the extent of the challenge, as not even a different modality is easy to detect without our proposed approach.  In practice, we evaluate three~criteria \xadded{according to which a decision of whether a data instance is an outlier can be made}:
\begin{enumerate}
\item The classifier's predictive entropy, as recently suggested to work surprisingly well in deep networks \cite{Hendrycks2017} but technically well known to be overconfident \cite{Matan1990}. \xadded{The intuition here is that the predictive entropy $- \sum_{y \in C} p(y|x) \log{p(y|x)}$ considers the probability of all other classes and is at a maximum if the distribution is uniform, i.e., when the confidence in the prediction is low.}
\item The generative model's obtained negative log-likelihood, to concur with previous findings \cite{Nalisnick2019, Ovadia2019} on overconfidence in generative models. On the basis of Equation \eqref{eq:general_loss}, the intuition is that the negative log-likelihood should be much larger for unseen data.
\item Our suggested OpenVAE aggregate posterior-based EVT approach, according to the outlier likelihood introduced Equation \eqref{eq:outlier_probability}.
\end{enumerate}

\subsubsection*{Results}
\begin{figure}[H]

\includegraphics[width = 0.3 \textwidth]{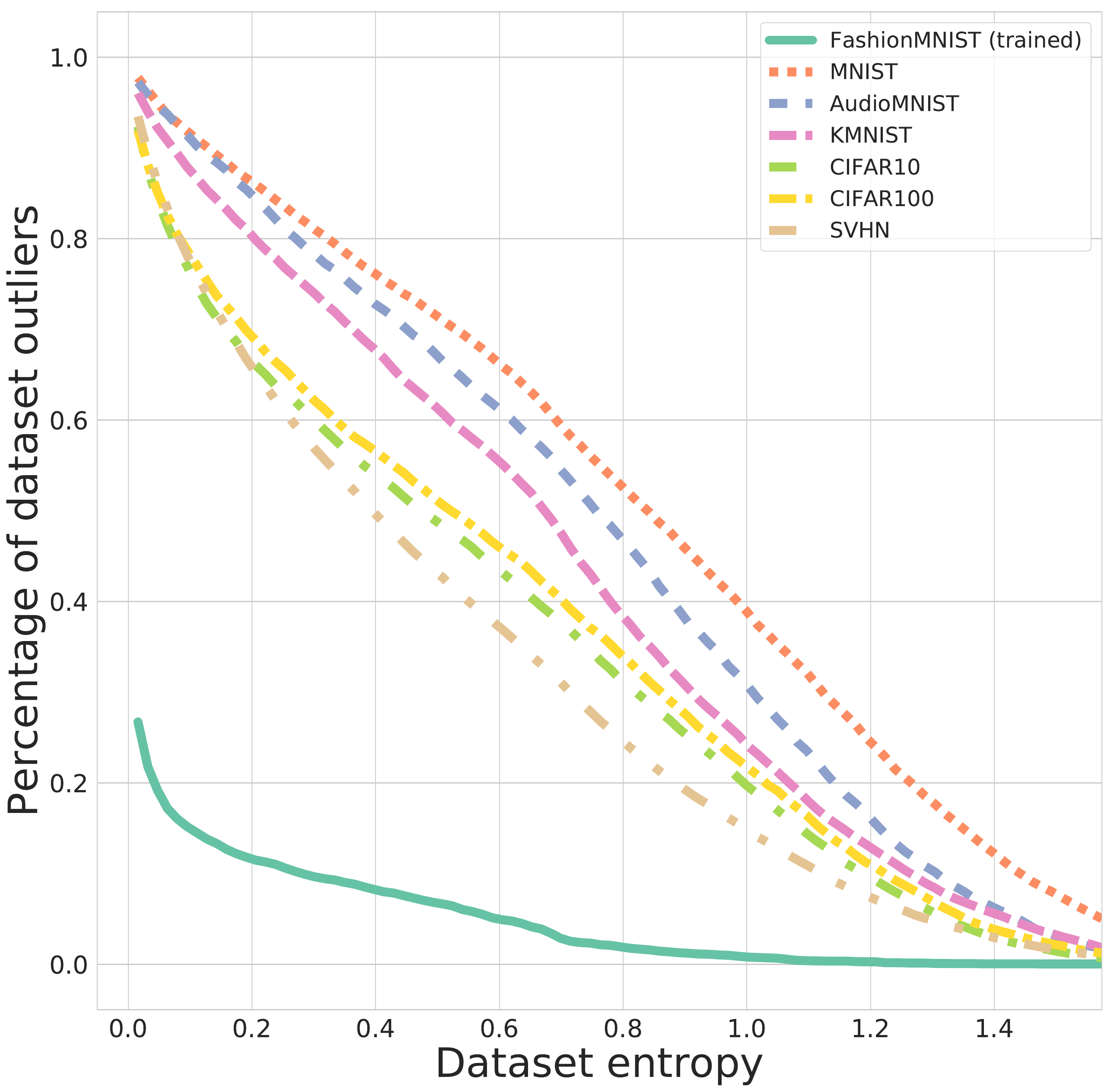}
\quad
\includegraphics[width = 0.3 \textwidth]{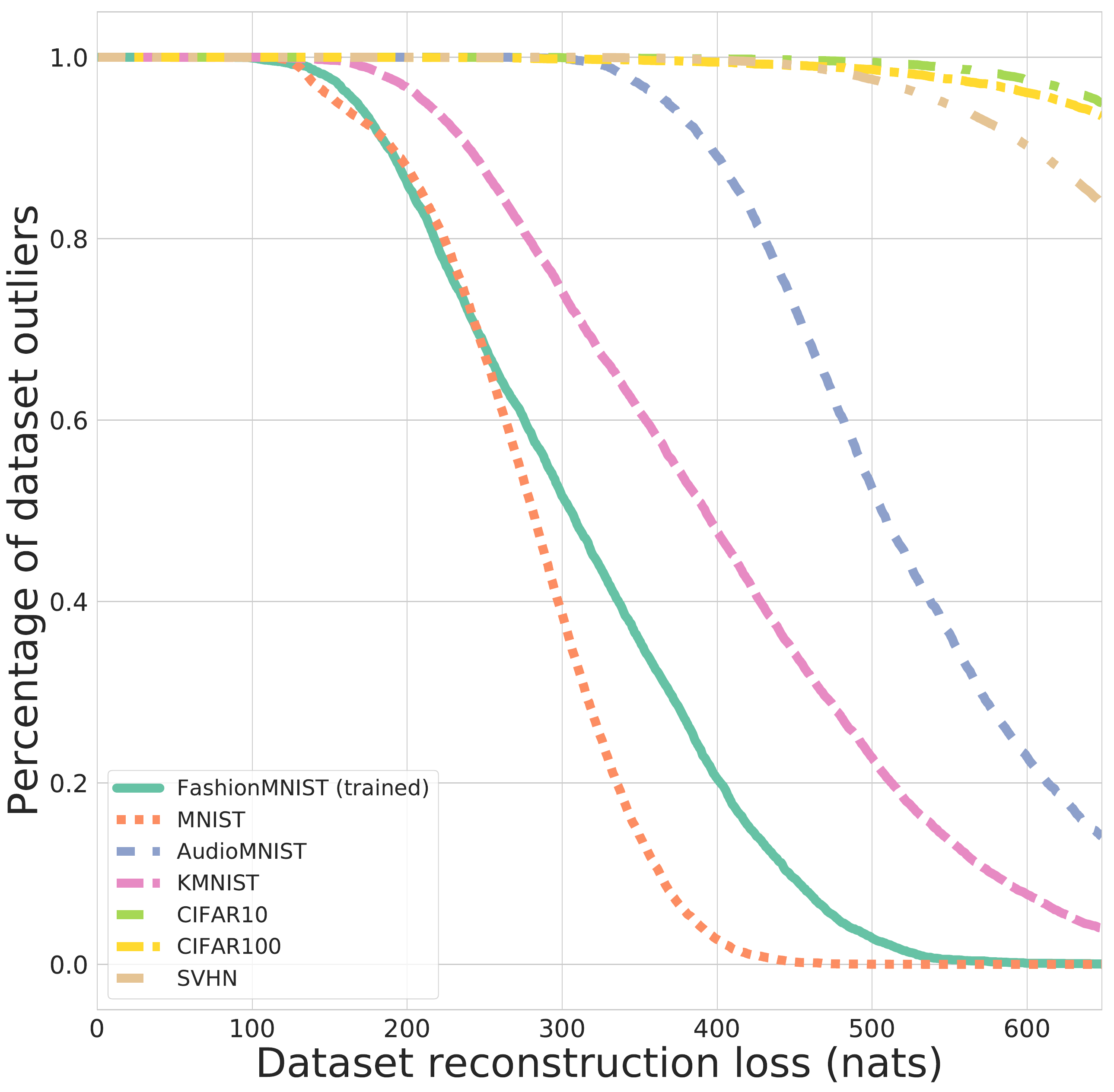}
\quad
\includegraphics[width = 0.3 \textwidth]{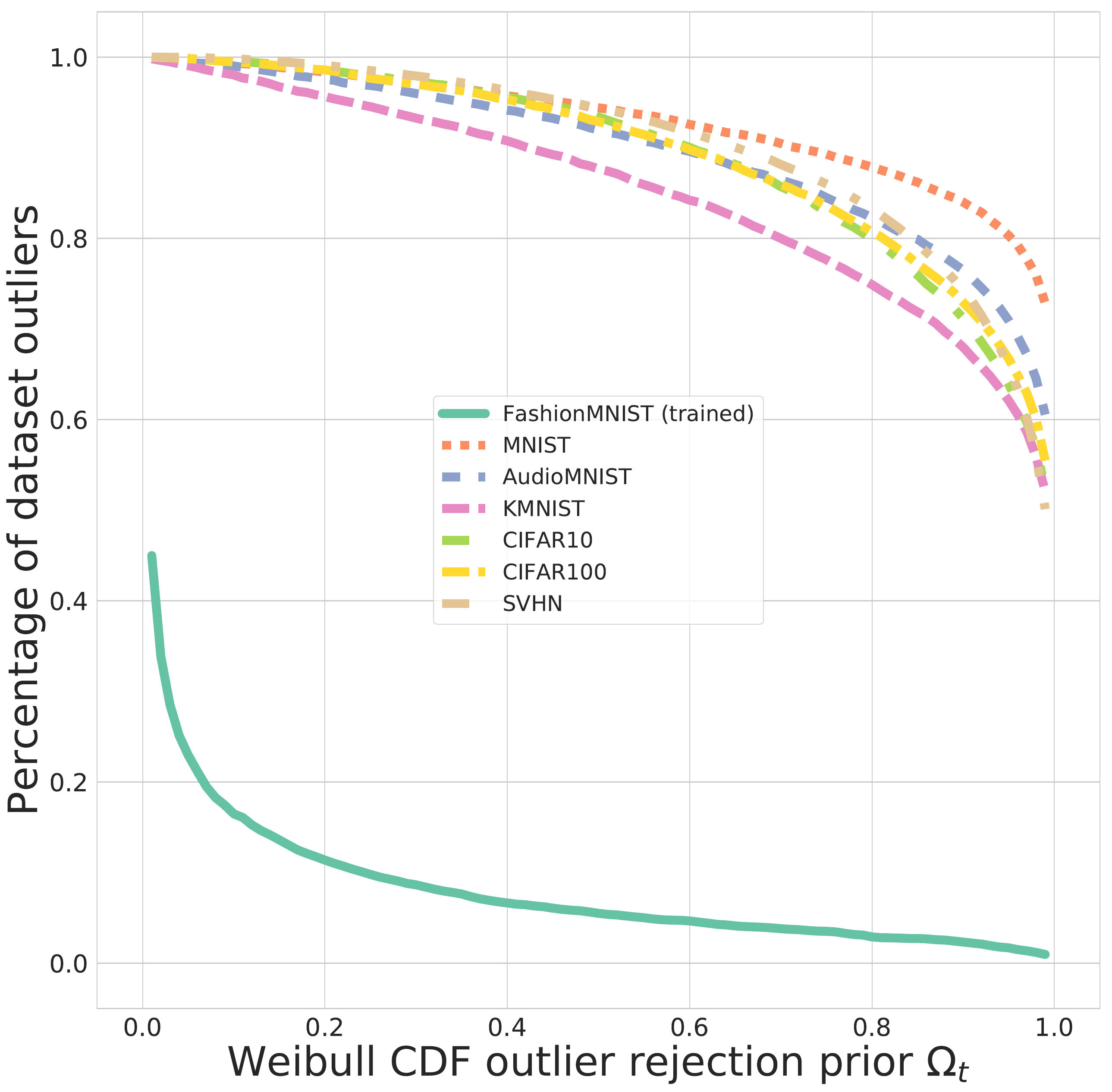}
\caption{Model trained on FashionMNIST evaluated on unknown datasets. \xadded{Robust classification of a known dataset (percentage of dataset outliers at 0\%), while correctly flagging unknown datasets as outlying (percentage of dataset outliers at 100\%), occurs when the solid green curve is separated from any of the colored dashed curves.} (\textbf{Left}) Classifier entropy is insufficient to separate unknown from the known task's test data. (\textbf{Center}) Reconstruction log-likelihood allows for a partial distinction. (\textbf{Right}) Our posterior-based EVT approach in OpenVAE considers the large majority of unknown data as statistical outliers across a wide range of rejection priors $\Omega_{t}$.}
\label{fig:outlier}
\end{figure}

Figure \ref{fig:outlier} provides a qualitative intuition behind the three criteria and respective percentage of the total dataset being considered as outlying for FashionMNIST. Consistent with \citet{Nalisnick2019}, we can observe that the use of reconstruction loss can sometimes distinguish between the known tasks' test data and unknown datasets but results in failure for others. In the case of the classifier predictive entropy, depending on the exact choice of entropy threshold, generally only a partial separation can be achieved. Furthermore, both of these criteria pose the additional challenge of the  results being highly dependent on the choice of the precise cut-off value. In contrast, the test data from the known tasks is regarded as inlying across a wide range of rejection priors $\Omega_{t}$ for Equation \eqref{eq:outlier_probability}, and the majority of other datasets is consistently regarded as outlying by our introduced OpenVAE~approach.

Corresponding quantitative outlier detection accuracies are provided in Table \ref{tab:openset_results}. To find thresholds for the sensitive entropy and reconstruction curves, we used a $5\%$ validation split to determine the respective value at which $95\%$ of the validation data is considered as inlying before using these priors to determine outlier counts for the known tasks' test set as well as other datasets. \xadded{In an intuitive picture, we ``trace'' the solid green curve of Figure~\ref{fig:outlier} for a validation set of the originally trained dataset, check where we intersect with the $x$-axis for a $y$-axis value of 5\% and then fix the corresponding criterion's value at this point as an outlier rejection threshold for testing. We then report the percentage of the test set being considered as an outlier, together with the percentage for various unknown datasets.} In the table, we \xadded{additionally} extend our intuition of Figure \ref{fig:outlier} to now further investigate what would happen if we had not trained a single VAE model that learned reconstruction and classification according to Equation (\ref{eq:general_loss}) but separate models.  For this purpose,  we also investigate a dual model approach, i.e., a purely discriminative deep-neural-network-based classifier and a separate unsupervised VAE (Equation \eqref{eq:general_loss} without blue terms).

In this way, we can showcase the advantages of a generative modeling formulation that considers the joint distribution $p(\boldsymbol{x},\boldsymbol{y})$ in conjunction with EVT.
For instance, we can compare our values with the purely discriminative OpenMax EVT approach \cite{Bendale2015}. At the same time, this provides a justification for why the existing continual-learning approaches of the next section, especially those relying on the maintenance of multiple models, are non-ideal, as they cannot seem to adequately solve the open-set challenge.



In terms of the obtained results, with the exception of MNIST, which appears to be an easy to identify dataset for all approaches, we can make two key observations:

\begin{enumerate}
\item Both EVT approaches generally outperform the other criteria, particularly for our suggested aggregate posterior-based OpenVAE variant, where a near perfect open-set detection can be achieved.
\item Even though EVT can be applied to purely discriminative models (as in OpenMax), the generative OpenVAE model trained with variational inference consistently exhibited more accurate outlier detection. We posit that this robustness is due to OpenVAE explicitly optimizing a variational lower bound that considers the data distribution $p(\boldsymbol{x})$ in addition to a pure optimization of features that maximize $p(y|\boldsymbol{x})$.
\end{enumerate}

\subsubsection*{Open Set Recognition with Monte-Carlo Dropout Based Uncertainty}


One might be tempted to assume that the trained weights of the individual deep neural network encoder layers are still deterministic and the failure of predictive entropy as a measure for unseen unknown data could thus primarily be attributed to uncertainty not being expressed adequately. Placing a distribution on the weights, akin to a fully Bayesian neural network, would then be expected to resolve this issue. For this purpose, we further repeat all of our experiments by treating the model weights as the random variable being marginalized through the use of Monte-Carlo Dropout (MCD) \cite{Gal2015}.~Accordingly,  the models were re-trained with a Dropout probability of $0.2$ in each layer. We then conducted 50~stochastic forward passes through the entire model for prediction. The obtained open-set recognition results are reported in Table \ref{tab:openset_results_MCD}.

Although MCD boosts the outlier detection accuracy, particularly for criteria, such as predictive entropy, the previous insights and drawn conclusions still hold. In summary, the joint generative model generally outperforms a purely discriminative model in terms of open-set recognition, independently of the used metric, and our proposed aggregate posterior-based EVT approach of OpenVAE yields an almost perfect separation of known and unseen unknown data. Interestingly, this was already  achieved in the prior table without MCD. Resorting to the repeated model calculation of MCD thus appears to be without enough of an advantage to warrant the added computational complexity in the context of posterior-based open-set recognition, a further key advantage of OpenVAE.

\subsection{Learning Classes Incrementally in Continual Learning}

To showcase how our OpenVAE approach mitigates catastrophic interference \xadded{in addition to successfully handling unknown data in robust prediction, we conduct an investigation of the test accuracy when learning classes incrementally.}

\subsubsection*{Experimental Set-Up and Evaluation}
We consider the incremental MNIST dataset (where classes arrive in groups of two) and the corresponding versions of the FashionMNIST and AudioMNIST datasets, similar to popular literature \cite{Zenke2017, Kirkpatrick2017, Farquhar2018, Shin2017, Parisi2019}.  We re-emphasize that such a setting has a sole focus on mitigating catastrophic interference and does not account for the the challenges presented in the previous open-set recognition section,  which we detail in the prospective discussion section.   For a flexible comparison, we report our aggregate posterior-based generative replay approach in OpenVAE on both a simple multi-layer perceptron (MLP), as well as a deep convolutional neural network (CNN) based on wide residual networks (WRN). For the former, we follow previous continual-learning studies and employ a two-hidden-layer and 400-unit multi-layer perceptron \cite{Kemker2018}. For the latter, we use both encoder and decoder architectures of 14-layer wide residual networks~\cite{Zagoruyko2016, He2016} with a latent dimensionality of 60 \cite{Gulrajani2017, Chen2016}. For our statistical outlier rejection, we use a  rejection prior of $\Omega_{t} = 0.01$ and dynamically set tail-sizes to 5\% of seen examples per class.

For our own experiments, we report the mean and standard deviation \xadded{of the average classification test accuracy} across five experimental repetitions.  If our re-implementation of related works  achieved a better than original value,  we report this number, otherwise the work that  reported the specific best value is cited next to it. The full training details, \xadded{including details on hardware and code,} are supplied in Appendix \ref{app:sec_hyperparams}.

\begin{table}[H]

\caption{Outlier detection values of the joint model and separate discriminative and generative models (denoted as ``CNN + VAE''; discriminative convolutional neural network and variational auto-encoder), when considering 95\% of the known tasks' validation data as inlying. The percentage of detected outliers is reported based on the classifier predictive entropy, reconstruction negative log-likelihood (NLL) and our posterior-based extreme-value theory approach. Note that larger values are better, except for the test data of the trained dataset, where ideally 0\% should be considered as outlying. \xadded{The outlier detection values have additionally been color coded, where \textit{worse} results appear in red. A deeper shading thus indicates a method's failure to robustly recognize unknown data as such. With this color coding, we can easily see how MNIST appears to be an easy to identify dataset for all approaches; however, we notice right away that our OpenVAE is the \textit{only} method (row) that does \textit{not} have a single red value for any dataset combination. In fact, the lowest outlier detection accuracy of OpenVAE is a very high 94.76\%.} }
\label{tab:openset_results}

\begin{adjustwidth}{-\extralength}{0cm}\setlength{\tabcolsep}{1.83mm}
{\small\begin{tabular}{llll*{7}{R}}
\toprule

\multicolumn{4}{c}{\textbf{Outlier Detection at 95\% Validation Inliers (\%)}} & \multicolumn{1}{c}{\textbf{MNIST}} & \multicolumn{1}{c}{\textbf{Fashion}} & \multicolumn{1}{c}{\textbf{Audio}} & \multicolumn{1}{c}{\textbf{KMNIST}} & \multicolumn{1}{c}{\textbf{CIFAR10}} & \multicolumn{1}{c}{\textbf{CIFAR100}} & \multicolumn{1}{c}{\textbf{SVHN}} \\
\textbf{Trained} & \textbf{Model} & \textbf{Test Acc.} & \textbf{Criterion} & \multicolumn{1}{c}{} & \multicolumn{1}{c}{} & \multicolumn{1}{c}{} & \multicolumn{1}{c}{} & \multicolumn{1}{c}{} & \multicolumn{1}{c}{} & \multicolumn{1}{c}{} \\
\hline

\multirow{7}{*}{\STAB{\rotatebox[origin=c]{90}{\textbf{MNIST}}}} & Dual, & 99.40 & Class entropy & \multicolumn{1}{c}{4.160} & 90.43 & 97.53 & 95.29 & 98.54 & 98.63 & 95.51 \\
& CNN +& & Reconstruction NLL & \multicolumn{1}{c}{5.522} & 99.98 & 99.97 & 99.98 & 99.99 & 99.96 & 99.98 \\
& VAE & & OpenMax & \multicolumn{1}{c}{4.362} & 99.41 & 99.80 & 99.86 & 99.95 & 99.97 & 99.52 \\
\cmidrule{2-11}
& Joint & 99.53 & Class entropy  & \multicolumn{1}{c}{3.948} & 95.15 & 98.55 & 95.49 & 99.47 & 99.34 & 97.98 \\
& VAE & & Reconstruction NLL & \multicolumn{1}{c}{5.083} & 99.50 & 99.98 & 99.91 & 99.97 & 99.99 & 99.98 \\
& & & OpenVAE (ours) & \multicolumn{1}{c}{4.361} & 99.78 & 99.67 & 99.73 & 99.96 & 99.93 & 99.70 \\
\hline
\multirow{6.5}{*}{\STAB{\rotatebox[origin=c]{90}{\textbf{FashionMNIST}}}} & Dual, & 90.48 & Class entropy & 74.71 & \multicolumn{1}{c}{5.461} & 69.65 & 77.85 & 24.91 & 28.76 & 36.64 \\
& CNN + & & Reconstruction NLL & 5.535 & \multicolumn{1}{c}{5.340} & 64.10 & 31.33 & 99.50 & 98.41 & 97.24 \\
& VAE & & OpenMax & 96.22 & \multicolumn{1}{c}{5.138} & 93.00 & 91.51 & 71.82 & 72.08 & 73.85 \\
\hhline{~----------}
& Joint & 90.92 & Class Entropy & 66.91 & \multicolumn{1}{c}{5.145} & 61.86 & 56.14 &  43.98 & 46.59 & 37.85 \\
& VAE & & Reconstruction NLL & 0.601 & \multicolumn{1}{c}{5.483} & 63.00 & 28.69 & 99.67 & 98.91 & 98.56 \\
& & & OpenVAE (ours) & 96.23 & \multicolumn{1}{c}{5.216} & 94.76 & 96.07 & 96.15 & 95.94 & 96.84 \\
\hline
\multirow{7}{*}{\STAB{\rotatebox[origin=c]{90}{\textbf{AudioMNIST}}}} & Dual, & 98.53 & Class entropy & 97.63 & 57.64 & \multicolumn{1}{c}{5.066} & 95.53 & 66.49 & 65.25 & 54.91 \\
& CNN + & & Reconstruction NLL & 6.235 & 46.32 & \multicolumn{1}{c}{4.433} & 98.73 & 98.63 & 98.63 & 97.45 \\
& VAE & & OpenMax & 99.82 & 78.74 & \multicolumn{1}{c}{5.038} & 99.47 & 93.44 & 92.76 & 88.73 \\
\hhline{~----------}
& Joint & 98.57 & Class entropy & 99.23 & 89.33 & \multicolumn{1}{c}{5.731} & 99.15 & 92.31 & 91.06 & 85.77 \\
& VAE & & Reconstruction NLL & 0.614 & 38.50 & \multicolumn{1}{c}{3.966} & 36.05 & 98.62 & 98.54 & 96.99 \\
& & & OpenVAE (ours) & 99.91 & 99.53 & \multicolumn{1}{c}{5.089} & 99.81 & 100.0 & 99.99 & 99.98 \\
\bottomrule
\end{tabular} }
\end{adjustwidth}
\end{table}

\begin{table}[H]
\centering
\caption{Outlier detection values of the joint model and separate discriminative and generative models (denoted as ``CNN + VAE''; discriminative convolutional neural network and variational auto-encoder), when considering 95\% of known tasks’ validation data is inlying. The percentage of detected outliers is reported based on classifier predictive entropy, reconstruction negative log-likelihood (NLL) and our posterior-based EVT approach. In contrast to Table \ref{tab:openset_results}, the results are now averaged over 50 Monte-Carlo dropout samples, with $p_{dropout} = 0.2$ for each layer, per data-point, respectively, to assess the model uncertainty. Note that larger values are better, except for the test data of the trained dataset, where ideally 0\% should be considered as outlying. \xadded{The color coding is analogous to Table \ref{tab:openset_results}.}}
\label{tab:openset_results_MCD}

\begin{adjustwidth}{-\extralength}{0cm}\setlength{\tabcolsep}{1.83mm}
{\small\begin{tabular}{llll*{7}{R}}\toprule

\multicolumn{4}{c}{\textbf{Outlier Detection at 95\% Validation Inliers (\%)}} & \multicolumn{1}{c}{\textbf{MNIST}} & \multicolumn{1}{c}{\textbf{Fashion}} & \multicolumn{1}{c}{\textbf{Audio}} & \multicolumn{1}{c}{\textbf{KMNIST}} & \multicolumn{1}{c}{\textbf{CIFAR10}} & \multicolumn{1}{c}{\textbf{CIFAR100}} & \multicolumn{1}{c}{\textbf{SVHN}} \\
\textbf{Trained} & \textbf{Model} & \textbf{Test Acc.} & \textbf{Criterion} & \multicolumn{1}{c}{} & \multicolumn{1}{c}{} & \multicolumn{1}{c}{} & \multicolumn{1}{c}{} & \multicolumn{1}{c}{} & \multicolumn{1}{c}{} & \multicolumn{1}{c}{} \\
\hline
\multirow{6.5}{*}{\STAB{\rotatebox[origin=c]{90}{\textbf{MNIST}}}} & Dual, & 99.41 & Class entropy & \multicolumn{1}{c}{4.276} & 91.88 & 96.50 & 96.65 & 95.84 & 97.37 & 98.58 \\
& CNN +& & Reconstruction & \multicolumn{1}{c}{4.829} & 99.99 & 100.0 & 99.90 & 100.0 & 100.0 & 100.0 \\
& VAE & & OpenMax & \multicolumn{1}{c}{4.088} & 87.84 & 98.06 & 95.79 & 97.34 & 98.30 & 95.74 \\
\hhline{~----------}
& Joint, & 99.54 & Class entropy  & \multicolumn{1}{c}{4.801} & 97.63 & 99.38 & 98.01 & 99.16 & 99.39 & 98.90 \\
& VAE & & Reconstruction & \multicolumn{1}{c}{5.264} & 99.98 & 100.0 & 100.0 & 100.0 & 100.0 & 100.0 \\
& & & OpenVAE (ours) & \multicolumn{1}{c}{4.978} & 99.99 & 100.0 & 99.94 & 99.96 & 99.95 & 99.68 \\
\hline

\multirow{6.5}{*}{\STAB{\rotatebox[origin=c]{90}{\textbf{FashionMNIST}}}} & Dual, & 90.58 & Class entropy & 75.50 & \multicolumn{1}{c}{5.366} & 70.78 & 74.41 & 49.42 & 49.17 & 38.84 \\
& CNN + & & Reconstruction NLL & 55.45 & \multicolumn{1}{c}{5.048} & 59.99 & 99.83 & 99.35 & 99.35 & 99.62 \\
& VAE & & OpenMax & 77.03 & \multicolumn{1}{c}{4.920} & 55.48 & 70.23 & 58.73 & 57.06 & 44.54 \\
\hhline{~----------}
& Joint, & 91.50 & Class Entropy & 85.05 & \multicolumn{1}{c}{4.740} & 67.90 & 78.04 & 63.89 & 66.11 & 59.42 \\
& AE & & Reconstruction & 1.227 & \multicolumn{1}{c}{5.422} & 85.85 & 39.76 & 99.94 & 99.72 & 99.99 \\
& & & OpenVAE (ours) & 95.83 & \multicolumn{1}{c}{4.516} & 94.56 & 96.04 & 96.81 & 96.66 & 96.28 \\
\hline
\multirow{7}{*}{\STAB{\rotatebox[origin=c]{90}{\textbf{AudioMNIST}}}} & Dual, & 98.76 & Class entropy & 99.97 & 61.26 & \multicolumn{1}{c}{4.996} & 96.77 & 63.78 & 65.76 & 59.38 \\
& CNN + & & Reconstruction NLL & 7.334 & 52.37 & \multicolumn{1}{c}{5.100} & 98.19 & 99.97 & 99.90 & 99.96 \\
& VAE & & OpenMax & 92.74 & 67.18 & \multicolumn{1}{c}{5.073} & 90.41 & 90.56 & 90.97 & 89.58 \\
\hhline{~----------}
& Joint, & 98.85 & Class entropy & 99.39 & 89.50 & \multicolumn{1}{c}{5.333} & 99.16 & 94.66 & 95.12 & 97.13 \\
& VAE & & Reconstruction NLL & 15.81 & 53.83 & \multicolumn{1}{c}{4.837} & 41.89 & 99.90 & 99.82 & 99.95 \\
& & & OpenVAE (ours) & 99.50 & 99.27 & \multicolumn{1}{c}{5.136} & 99.75 & 99.71 & 99.59 & 99.91 \\
\bottomrule
\end{tabular} }
\end{adjustwidth}
\end{table}

\subsubsection*{Results} In Table \ref{tab:incremental_results}, we report the final accuracy after having trained on each of the five increments. For an overall reference, we provide the achievable upper-bound continual-learning performance, i.e.,~accumulating all data over time and optimizing Equation \eqref{eq:general_loss}. We can observe that our proposed OpenVAE approach provides significant improvement over generative replay with a conventional supervised VAE. In comparison with the immediately related works, our approach surpasses variational continual learning (VCL) \cite{Nguyen2018}, an approach that employs a full Bayesian neural network (BNN), with the additional benefit that our approach scales trivially to complex network architectures.

In contrast to variational generative replay (VGR) \cite{Farquhar2018}, OpenVAE initially appears to fall short. This is not surprising as VGR trains a separate GAN on each task's aggregate posterior, an apples to oranges comparison considering that we only use a single model. Nevertheless, even in a single model, we can surpass the multi-model VGR by leveraging recent advancements in generative modeling, e.g., by making the neural architecture more complex or augmenting our decoder with autoregressive sampling \cite{Chen2016, Gulrajani2017} (a complementary technique to OpenVAE, often also called PixelVAE and summarized in Appendix \ref{app:sec_gen_advances}).

At the bottom of Table \ref{tab:incremental_results}, we can see that this significantly improves upon the previously obtained accuracy.  The full accuracies, along with other metrics per dataset for all intermediate steps can be found in Appendix \ref{app:full_results}.

\begin{table}[H]
\caption{The accuracy $\alpha_{T}$ at the end of the last increment {\emph{T}} = 5 for class incremental learning approaches averaged over five runs. \xadded{For a fair comparison, if our re-implementation of related works  achieved a better than original value, we report our number, otherwise the work that  reported the specific best value is cited right next to the result.} Intermediate results can be found in Appendix \ref{app:full_results}. \label{tab:incremental_results}}
\vspace{-8pt}
\setlength{\tabcolsep}{6.48mm}
{\small \begin{tabular}{lccc}
\toprule
& \multicolumn{3}{c}{\textbf{\boldmath{Final Accuracy $\alpha_{T} (T=5)$ [\%]}}} \\
\midrule
\textbf{Method} &  \textbf{MNIST}  & \textbf{FashionMNIST}  & \textbf{AudioMNIST} \\
\midrule

MLP upper bound & 98.84 & 87.35 &  96.43\\
WRN upper bound & 99.29 & 89.24 &  97.87\\
\midrule
EWC \cite{Kirkpatrick2017} & 55.80 \cite{Chaudhry2018} & 24.48 $\pm$ $\tiny{2.86}$ & 20.48  $\pm$ $\tiny{1.73}$\\
DGR \cite{Shin2017} & 75.47 \cite{Hu2019} & 63.21 $\pm$ $\tiny{1.96}$ & 48.42  $\pm$ $\tiny{2.81}$\\
VCL \cite{Nguyen2018} & 72.30 \cite{Farquhar2018a} & 32.60 \cite{Farquhar2018a} & -\\
VGR \cite{Farquhar2018a} & 92.22 \cite{Farquhar2018a} & 79.10 \cite{Farquhar2018a} & - \\
Supervised VAE & 60.88 $\pm$ $\tiny{3.31}$ & 62.72 $\pm$ $\tiny{1.38}$ & 69.76  $\pm$ $\tiny{1.37}$\\
\midrule
OpenVAE---MLP & 87.31 $\pm$ $\tiny{1.22}$ & 66.14 $\pm$ $\tiny{0.50}$ & 81.84  $\pm$ $\tiny{1.44}$\\
OpenVAE---WRN & 93.24 $\pm$ $\tiny{3.74}$ & 69.88 $\pm$ $\tiny{1.71}$ & 87.72  $\pm$ $\tiny{1.59}$\\
OpenPixelVAE & 96.84 $\pm$ $\tiny{0.35}$ & 80.85 $\pm$ $\tiny{0.72}$ & 90.23  $\pm$ $\tiny{1.14}$\\\bottomrule
\end{tabular}}
\end{table}

\subsubsection*{High-Resolution Flower Images}

While the main goal of this paper is not to push the achievable boundaries of generation, we take this argument one step further and provide empirical evidence that our suggested aggregate posterior-based EVT sampling provides similar benefits when scaling to higher resolution color images. For this purpose, we consider the additional flowers dataset \cite{Nilsback2006} at a resolution of $256 \times 256$, investigated with five classes and increments of one class per step \cite{Wu2018, Zhai2019}.

\begin{figure}[H]
\includegraphics[width=0.8 \columnwidth]{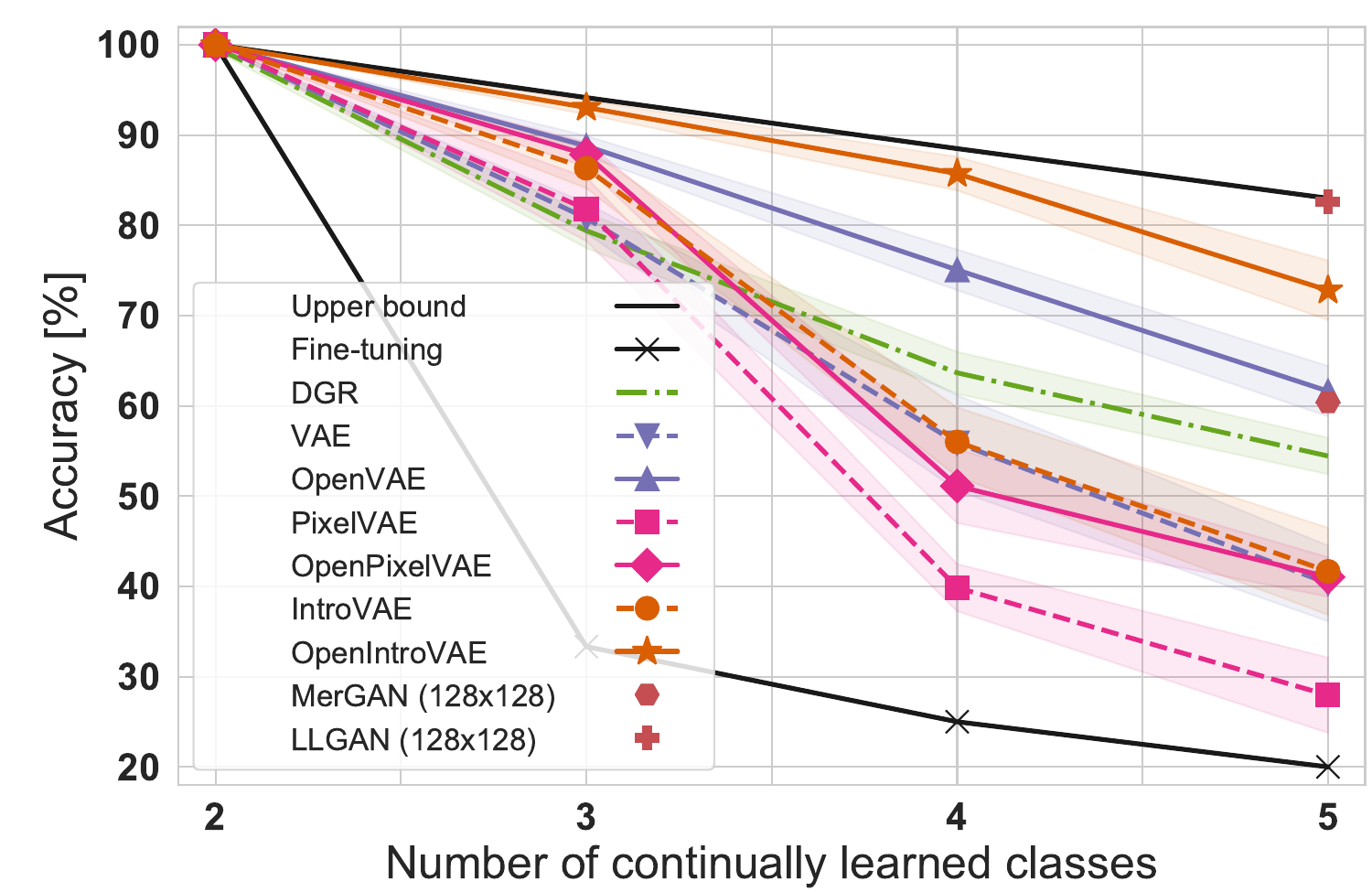}
\caption{Classification accuracy over five runs for continually learned flowers at $256 \times 256$ resolution to demonstrate how generative modeling advances draw similar benefits from our proposed aggregate posterior constrained generative replay (solid lines) over the open-set-unaware baselines (dashed counterparts).   \label{fig:flowers_accuracy}}
\end{figure}

In addition to autoregressive sampling, we also include a second complementary generative modeling improvement here, called VAEs with introspection (IntroVAE)~\cite{Huang2018}. A technical description of PixelVAE and IntroVAE is detailed in \mbox{Appendix \ref{app:sec_gen_advances}}.  For each generative modeling variant, including autoregression and introspection, we report the degradation of accuracy over time in Figure \ref{fig:flowers_accuracy} and demonstrate how their respective open-set-aware version provides substantial improvements. Intuitively, this improvement is due to an increase in the visual generation quality; see the examples in the earlier Figure \ref{fig:generated_openset}.



First, it is apparent how every OpenVAE variant improves upon its non open-set aware counterpart. We  further observe that the best version, OpenIntroVAE, appears to be in the same ballpark as  complex recent GAN approaches \cite{Wu2018, Zhai2019}, even though they do not solve the open-set recognition challenge and conduct a simplified evaluation. The latter works use a lower resolution of $128 \times 128$ (we were unable to scale to satisfying results at higher resolution) with additional distillation mechanisms, a continuously trained generator but a classifier that is trained and assessed only once at the end. We nevertheless report the respective values for intuition. We conclude that the obtained final accuracy can be competitive and is remarkably close to the achievable upper bound. A suspected initial VAEs generation quality limitation appears to be lifted with modern extensions and our proposed sampling scheme.

\begin{figure}[H]

\includegraphics[width= 0.95 \textwidth]{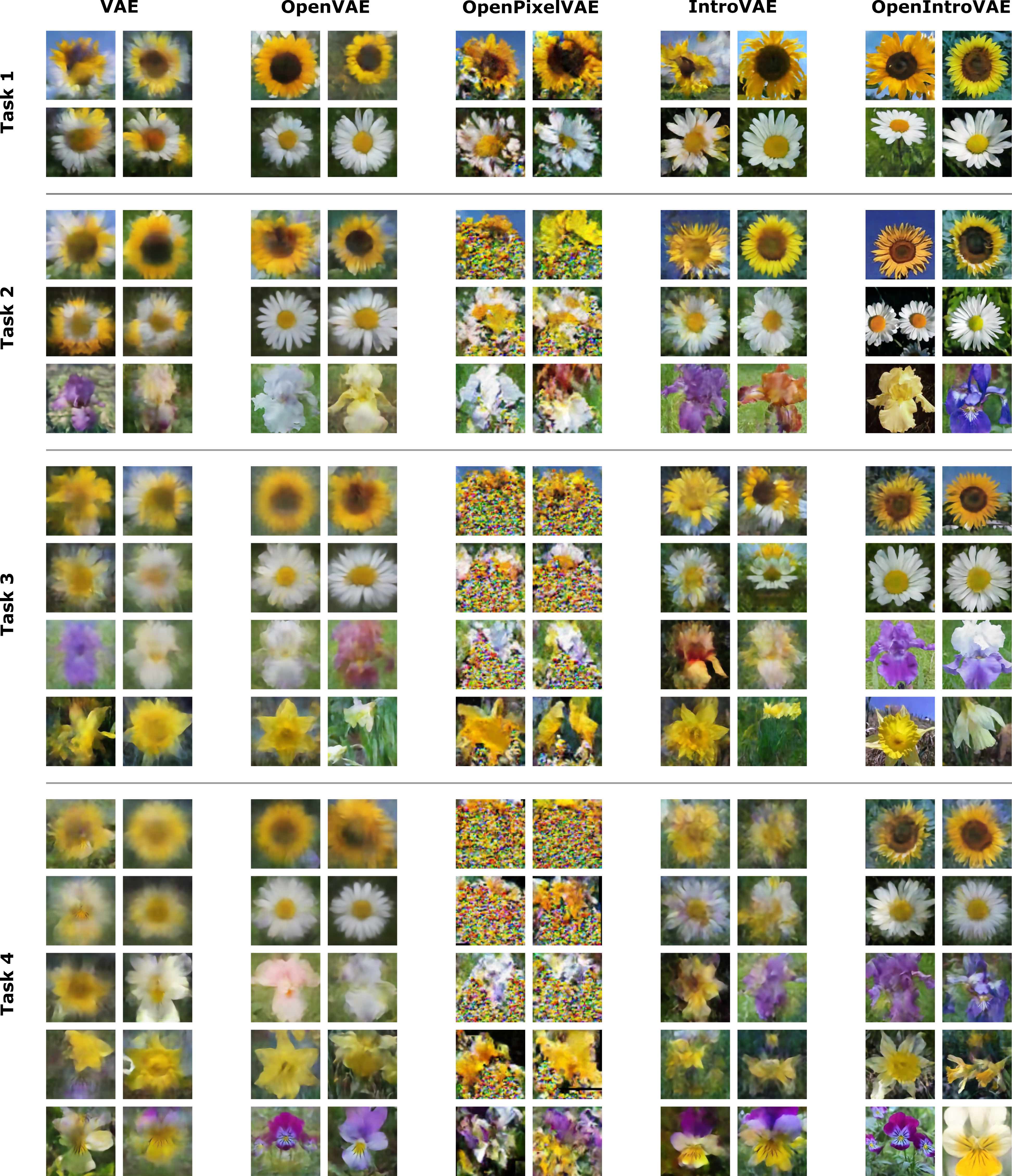}
\caption{Generated $256 \times 256$ flower images for various continually trained models. Images were selected to provide a qualitative intuition behind the quantitative results of Figure \ref{fig:flowers_accuracy}.  Images are compressed for a side-by-side view.}\label{fig:flowers_comparison}
\end{figure}

We also support our quantitative statements visually with a few selected generated images for the various generative variants in Figure \ref{fig:flowers_comparison}.  We emphasize that these examples are supposed to primarily provide visual intuition in support of the existing quantitative results, as it is difficult to draw conclusions from a perceived subjective quality from a few images alone. From a qualitative viewpoint, the OpenVAE without generative modeling extensions appears to suffer from the limitations of a traditional VAE and generates blurry~images.

However, our open-set approach nevertheless provides a clearer disambiguation of classes, particularly already at the stage of task 2. The addition of introspection significantly increases the image detail, albeit still degrades considerably due to ambiguous interpolations in samples from low-density areas outside the aggregate posterior. This is again resolved by combining introspection with our proposed posterior-based EVT approach, where image quality is retained across multiple generative replay steps. From a purely visual perspective it is clear why this model outperforms the other approaches significantly in terms of quantitative accuracy values.

Interestingly, our visual inspection also hints at why the PixelVAE and its open-set variant perform much worse than perhaps initially expected. As the caveat is the same in both PixelVAE and OpenPixelVAE, we only show generated instances for the latter. From these samples, we can hypothesize why the initial performance is competitive but rapidly declines. It appears that the autoregression suffers from forgetting in terms of its long-range pixel dependency.

Whereas at the beginning, the information is locally consistent across the entire image, in each consecutive step, a further portion of subsequent pixels for old tasks is progressively replaced with uncorrelated noise. The conditioning thus appears to primarily be captured on new tasks only, resulting in interference effects.  We continue this discussion alongside potential other general limitations of generative modeling variant choices in Appendix \ref{app:sec_limitations}.

\vspace{-9pt}
\section{Discussion}

\xadded{As a final piece of discussion, we would like to recall and emphasize a few important points of how our results should be interpreted and contextualized.}

\subsection{Presence of Unknown Data and Current Benchmarks}

\xadded{Perhaps most importantly, we re-iterate that OpenVAE is unique in that it provides a grounded basis to conduct continual learning in the presence of unknown data.
However,  as evidenced from the quantitative open-set recognition results,  the inclusion of unknown data instances into continual learning would immediately result in the failure of the present continual-learning approaches at this point, simply because they lack a principled mechanism to provide robust predictions. For this reason, we show traditional incremental classification results as a proxy to assess our improved aggregate posterior-based generation~quality.

Our class incremental accuracy reports in this paper should thus be interpreted with caution as they represent only a part of OpenVAE's capability, similar to a typical ablation study. We nevertheless provided this type of comparison, in order to situate OpenVAE with respect to some existing generative continual-learning methods in terms of catastrophic forgetting, rather than presenting OpenVAE in isolation in a more realistic new setting.}

\subsection{State of the Art in Class Incremental Learning and Exemplar Rehearsal}

\xadded{Following the above subsection, we note that a fair comparison of realistic class incremental learning is further complicated due to various involved factors. In fact, multiple related works make various additional assumptions on the extra storage of explicit data subsets and the use of multiple generative models per task or even multiple classifiers. We do not make these assumptions here in favor of generality. In this spirit, we focused our evaluation on our contributions' relevant novelty with respect to combining the detection of unknown data with the prevention of catastrophic forgetting in generative models.

The introduced OpenVAE shows that both are achievable simultaneously. At the same time, the reader familiar with the recent continual-learning literature will likely notice that some modern approaches that are attributed with state of the art in class incremental learning have not been included in our comparison. These approaches all fall into the category of exemplar rehearsal. We would like to emphasize that this is deliberate and not out of ignorance, as we see these works as purely complementary. We nevertheless wish to give deserved credit to these works and provide an outlook to one future research~direction.}

\xadded{The primary reason for omitting a direct comparison with state of the art works in continual learning that employ exemplar rehearsal is that we believe such a comparison would be misleading. In fact, contrasting our OpenVAE against these works would imply that these methods are somehow competing. In reality, exemplar rehearsal, or the so called extraction of core sets, is an auxiliary mechanism that can be applied out-of-the-box to our experimental set-up in this work. The main premise here is that catastrophic forgetting in continual learning can be reduced by retaining an explicit subset of the original data and subsequently continuously interleaving this stored data into the training process.

Early works, such as iCarl \cite{Rebuffi2017} show that performance is then a function of two key aspects: the data selection technique and the memory buffer size. The former, selection of an appropriate data subset, essentially boils down to a non-continual-learning question, i.e., how to approximate the entire distribution through only a few instances. Exemplar rehearsal works thus make use of existing techniques here, such as core sets \cite{Bachem2015}, herding~\cite{Welling2009}, nearest mean-classifiers \cite{Mensink2012} or simply picking data samples uniformly at random~\cite{Prabhu2020}.

The second question, on memory buffer size, has an almost trivial answer. The larger the memory buffer size, the better the performance. This is intuitive, yet also makes comparison challenging, as a memory buffer of the size of the entire dataset is analogous to what we referred to as ``incremental upper bound'' in our experiments. If we were to simply store the complete dataset, then catastrophic forgetting would be avoided entirely. Modern class incremental learning works make heavy use of this fact and store large portions of the original data, showing that the more data is stored, the higher the performance goes.

Primary examples include the recent works on Mnemonics Training~\cite{Liu2020}, Contrastive Continual Learning (Co2L) \cite{Cha2021} or Dark Experience Replay (DER) \cite{Buzzega2020}. We do not wish to dive into a discussion here of whether or not such data storage is realistic or what size of a memory buffer should be assumed. A respective reference that questions and discusses whether storing of original data is synonymous with progress in continual learning is Greedy Sampler and Dumb Learner (GDumb) \cite{Prabhu2020}, where it is shown that the amount of extracted data alone amounts to a significant portion of ``state-of-the-art'' performance.

Primarily, we point out that the latter works all show that a larger memory buffer shows ``better'' class incremental learning performance, i.e., less forgetting. However, most importantly, extracting and storing parts of the original data into a separate memory buffer is an auxiliary process that is entirely complementary to our propositions of OpenVAE. As such, each of the methods referenced in this subjection is straightforward to combine with our work. Although we see such a combination as important prospective work, we leave detailed experimentation up to future investigations.

The rationale behind this choice is that inclusion of a memory buffer will inevitably additionally boost the performances of the results of Table \ref{tab:incremental_results}, yet provide no additional insights to our main hypothesis and contribution: the proposition of OpenVAE to show that detection of unknown data for robust prediction can effectively be achieved alongside reduction of catastrophic forgetting in continual learning.}

\section{Conclusions}
We proposed an approach to unify the prevention of catastrophic interference in continual learning with open-set recognition based on variational inference in deep generative models. As a common denominator, we introduced EVT-based bounds to the aggregate posterior.  The correspondingly named OpenVAE was shown to achieve compelling results in being able to distinguish known from unknown data, while boosting the generation quality in continual learning with generative replay.

\xdeleted{As a final discussion, we emphasize that OpenVAE is unique in that it provides a grounded basis to conduct continual learning in the presence of unknown data.
However,  as evident from the quantitative open-set recognition results,  inclusion of unknown data instances into continual learning would immediately result in failure of present continual-learning approaches at this point, simply because they lack a principled mechanism to provide robust predictions.  For this reason, we demonstrated traditional incremental classification results  as a proxy to assess our improved aggregate posterior-based generation quality.  Our class incremental accuracy reports in this paper should thus be interpreted with caution, as they represent only a part of OpenVAE's capability,  similar to a typical ablation study. We nevertheless provided this type of comparison, in order to situate OpenVAE with respect to existing continual-learning methods in terms of catastrophic forgetting, rather than presenting OpenVAE in isolation in a more realistic new setting. Note that a fair comparison is further complicated because multiple related works make various further assumptions on additional storage of explicit data subsets, use of multiple generative models per task or even multiple classifiers, which we do not make.  We leave the use of memory buffers of original data to alleviate catastrophic interference to future investigations, as such an approach is independent and entirely complementary to any employed continual-learning model. (moved)}

We believe that our demonstrated benefits from recent generative modeling techniques in the context of high-resolution flower images with OpenVAE provide a natural synergy to be explored in a range of future applications.  We envision prospective works to employ OpenVAE as a baseline when relaxing the closed-world assumption in continual learning and allowing unknown data to appear in the investigated benchmark streams at all times in the move to a more realistic evaluation.

\vspace{6pt}

\authorcontributions{The authors contributed to this work in the following ways: Conceptualization, M.M. and V.R.; methodology, M.M.; software, M.M., I.P. and S.M.; validation,  M.M., S.M. and Y.H.; formal analysis, M.M. and I.P.; investigation, M.M.; resources, V.R.; writing---original draft preparation, M.M. and I.P.; writing---review and editing, M.M.,  I.P. and V.R.; visualization,  M.M., I.P. and Y.H.; supervision, V.R.; project administration, M.M.; funding acquisition, V.R. All authors have read and agreed to the published version of the manuscript.}

\funding{We acknowledge the influence of various projects in shaping this paper. The projects include EU H2020 Project AEROBI Grant number 687384, EU H2020 Project RESIST Grant number 769066, BMBF project AISEL funding number 01IS19062. Additional financial support from Goethe University was instrumental in concluding the research.}

\institutionalreview{Not applicable.}

\informedconsent{Not applicable.}


\dataavailability{\xadded{All investigated datasets in this paper are publicly accessible popular benchmarks. Respective citations to the original works are provided in the main body upon first mention of each dataset. For convenience, we briefly list all datasets here and provide the public URL: Modified National Institute of Standards and Technology database (MNIST) \cite{LeCun1998} {(\url{http://yann.lecun.com/exdb/mnist/}, accessed on 22 January 2022),} FashionMNIST \cite{Xiao2017} {(\url{https://github.com/zalandoresearch/fashion-mnist}, accessed on 22 January 2022),} AudioMNIST \cite{Becker2018} {(\url{https://github.com/soerenab/AudioMNIST}, accessed on 22 January 2022),} Kuzushiji-MNIST (KMNIST) \cite{Clanuwat2018} ({\url{http://codh.rois.ac.jp/kmnist/index.html.en}, accessed on 22 January 2022),} Street View House Numbers (SVHN) \cite{Netzer2011} {(\url{http://ufldl.stanford.edu/housenumbers/}, accessed on 22 January 2022}), Canadian Institute for Advanced Research (CIFAR) datasets CIFAR10 \& CIFAR100 \cite{Krizhevsky2009} {(\url{https://www.cs.toronto.edu/~kriz/cifar.html}, accessed on 22 January 2022),} Oxford Flowers \cite{Nilsback2006} {(\url{https://www.robots.ox.ac.uk/~vgg/data/flowers/}, accessed on 22 January 2022).}}}

\conflictsofinterest{The authors declare no conflict of interest.}

\appendixtitles{no} 
\appendixstart
\appendix


\section[\appendixname~\thesection]{}
Our appendix provides further details for the material presented in the main body.  We first present more in-depth explanations and discussions on the introduced general concepts. At the end of the appendix, we then follow up with a full set of experimental results to complement the investigation of the experimental section.  Specifically, the structure is as follows:

\begin{itemize}[labelsep=20mm]
\item[\textit{Appendix~\ref{App_sec_lowerbound}}] Derivation of the lower-bound, Equation \eqref{eq:general_loss} of the main body.
\item[\textit{Appendix~\ref{sec:appx_beta}}] Extended discussion, qualitative and quantitative examples for the role of $\beta$.
\item[\textit{Appendix~\ref{app:sec_gen_advances}}] \textls[-5]{Description of generative model extensions: autoregression and introspection}.
\item[\textit{Appendix~\ref{app:sec_hyperparams}}] The full specification of the training procedure and hyper-parameters, including exact architecture definitions.
\item[\textit{Appendix~\ref{app:sec_limitations}}] Discussion of limitations.
\end{itemize}

This is followed by full sets of quantitative continual-learning results for all task increments, including reconstruction losses, in part Appendix \ref{app:full_results}.
\clearpage 

\appendixtitles{yes}
\subsection[\appendixname~\thesubsection]{Lower-Bound Derivation}\label{App_sec_lowerbound}
As mentioned in the main body of the paper, in supervised continual learning, we are confronted with a dataset $\boldsymbol{D} \equiv \left\{ \left( \boldsymbol{x}^{(n)}, y^{(n)} \right) \right\}_{n=1}^{N}$, consisting of $N$ pairs of data instances $\boldsymbol{x}^{(n)}$ and their corresponding labels $y^{(n)} \in \left\{ 1 \ldots C \right\}$ for $C$ classes.
We consider a problem scenario similar to the one introduced in ``Auto-Encoding Variational Bayes'' \cite{Kingma2013}, i.e.,~we assume that there exists a data generation process responsible for the creation of the labeled data given some random latent variable $\boldsymbol{z}$. For simplicity, we follow the authors' derivation for our model with the additional inclusion of data labels, however, without the $\beta$ term that is present in the main body. We point to the next section for a discussion of $\beta$.

Ideally, we would like to maximize $p(\boldsymbol{x, y}) = \int p(\boldsymbol{z}) p(\boldsymbol{x,y}|\boldsymbol{z}) d\boldsymbol{z}$, where the integral and the true posterior density
\begin{equation}\label{eq:true_posterior}
p(\boldsymbol{z} |\boldsymbol{x, y} ) = \frac{p(\boldsymbol{x, y} | \boldsymbol{z}) p(\boldsymbol{z})}{p(\boldsymbol{x, y})}
\end{equation}
are intractable. We thus follow the standard practice of using variational Bayesian inference and introducing an approximation to the posterior $q(\boldsymbol{z})$, for which we will specify the exact form later. Making use of the properties of logarithms and applying the above Bayes rule, we can now write:
\begin{adjustwidth}{-\extralength}{0cm}
\begin{equation}
\log p(\boldsymbol{x,y}) = \int q(\boldsymbol{z}) [ \log p(\boldsymbol{x, y} | \boldsymbol{z}) + \log p(\boldsymbol{z}) - \log p(\boldsymbol{z} | \boldsymbol{x, y}) + \log q(\boldsymbol{z}) - \log q(\boldsymbol{z}) ] d\boldsymbol{z},
\end{equation}
\end{adjustwidth}
as the left-hand side is independent of $\boldsymbol{z}$ and $\int q(\boldsymbol{z}) d\boldsymbol{z} = 1$. Using the definition of the Kullback--Leibler divergence (KLD) $\kld{q}{p} = - \int q(x) \log(p(x) / q(x))$ we can rewrite this as:
\begin{equation}
\log p(\boldsymbol{x,y}) - \kld{q(\boldsymbol{z})}{p(\boldsymbol{z} | \boldsymbol{x, y})} = \mathbb{E}_{q(\boldsymbol{z})} \left[ \log p(\boldsymbol{x, y} | \boldsymbol{z}) \right] - \kld{q(\boldsymbol{z})}{p(\boldsymbol{z})} \, .
\end{equation}

Here, the right hand side forms a variational lower-bound to the joint distribution $p(\boldsymbol{x,y})$, as the \emph{KL} divergence between the approximate and true posterior on the left hand side is strictly positive.

At this point, we make two choices that deviate from prior works that made use of labeled data in the context of generative models for semi-supervised learning \cite{Kingma2014}. We assume a factorization of the generative process of the form $p(\boldsymbol{x}, \boldsymbol{y}, \boldsymbol{z}) = p(\boldsymbol{x}|\boldsymbol{z})p(\boldsymbol{y}|\boldsymbol{z})p(\boldsymbol{z})$ and introduce a dependency of $q(\boldsymbol{z})$ on $\boldsymbol{x}$ but not explicitly on $\boldsymbol{y}$, i.e., $q(\boldsymbol{z} | \boldsymbol{x})$. In contrast to class-conditional generation, this dependency essentially assumes that all information about the label can be captured by the latent $\boldsymbol{z}$, and there is thus no additional benefit in explicitly providing the label when estimating the data likelihood $p(\boldsymbol{x} | \boldsymbol{z})$.

This is crucial as our probabilistic encoder should be able to predict labels without requiring it as input to our model, i.e., $q(\boldsymbol{z}|\boldsymbol{x})$ instead of the intuitive choice of $q(\boldsymbol{z}|\boldsymbol{x, y})$. However, we would like the label to nevertheless be directly inferable from the latent $\boldsymbol{z}$. In order for the latter to be achievable, we require the corresponding classifier that learns to predict $p(\boldsymbol{y}|\boldsymbol{z})$ to be linear in nature. This guarantees linear separability of the classes in latent space, which can in turn then be used  for open-set recognition and the generation of specific classes as shown in the main body.

\subsection[\appendixname~\thesubsection]{Further Discussion on the Role of {$\beta$}}\label{sec:appx_beta}
In the main body, the role of the $\beta$ term \cite{Higgins2017} in our model's loss function is summarized briefly. Here, we delve into further detail with qualitative and quantitative examples to support the arguments made by prior works \cite{Hoffman2016, Burgess2017}. To facilitate the discussion, we repeat Equation (1) of the main body:\vspace{-3pt}
\begin{equation}
\begin{aligned}
\mathcal{L}\left(\boldsymbol{x}^{(n)}, \boldsymbol{y}^{(n)}; \boldsymbol{\theta}, \boldsymbol{\phi}, \boldsymbol{\xi} \right) =& \, \mathbb{E}_{q_{\boldsymbol{\theta}}(\boldsymbol{z} | \boldsymbol{x}^{(n)})} \left[ \log{p_{\boldsymbol{\phi}}(\boldsymbol{x}^{(n)} | \boldsymbol{z})} + \log{p_{\boldsymbol{\xi}}(y^{(n)} | \boldsymbol{z})} \right] \\
& - \beta \kld{q_{\boldsymbol{\theta}}(\boldsymbol{z} | \boldsymbol{x}^{(n)})}{p(\boldsymbol{z})}
\end{aligned}
\end{equation}

The $\beta$ term weights the strength of the regularization by the prior through the Kullback-Leibler (\emph{KL}) divergence. The selection of this strength is necessary to control the information bottleneck of the latent space and regulate the effective latent encoding overlap. To repeat the main body and previous arguments by \cite{Hoffman2016,Burgess2017}: too large $\beta$ values (typically $>>1$) will result in a collapse of any structure present in the aggregate posterior. Too small $\beta$ values (typically $<< 1$) lead to the latent space being a lookup table. In either case, there is no meaningful information between the latents. This effect is particularly relevant to our objective of linear class separability, which requires the formation of an aggregate latent encoding that is disentangled with respect to the different classes.

\begin{figure}[H]

\includegraphics[width = 0.22 \columnwidth]{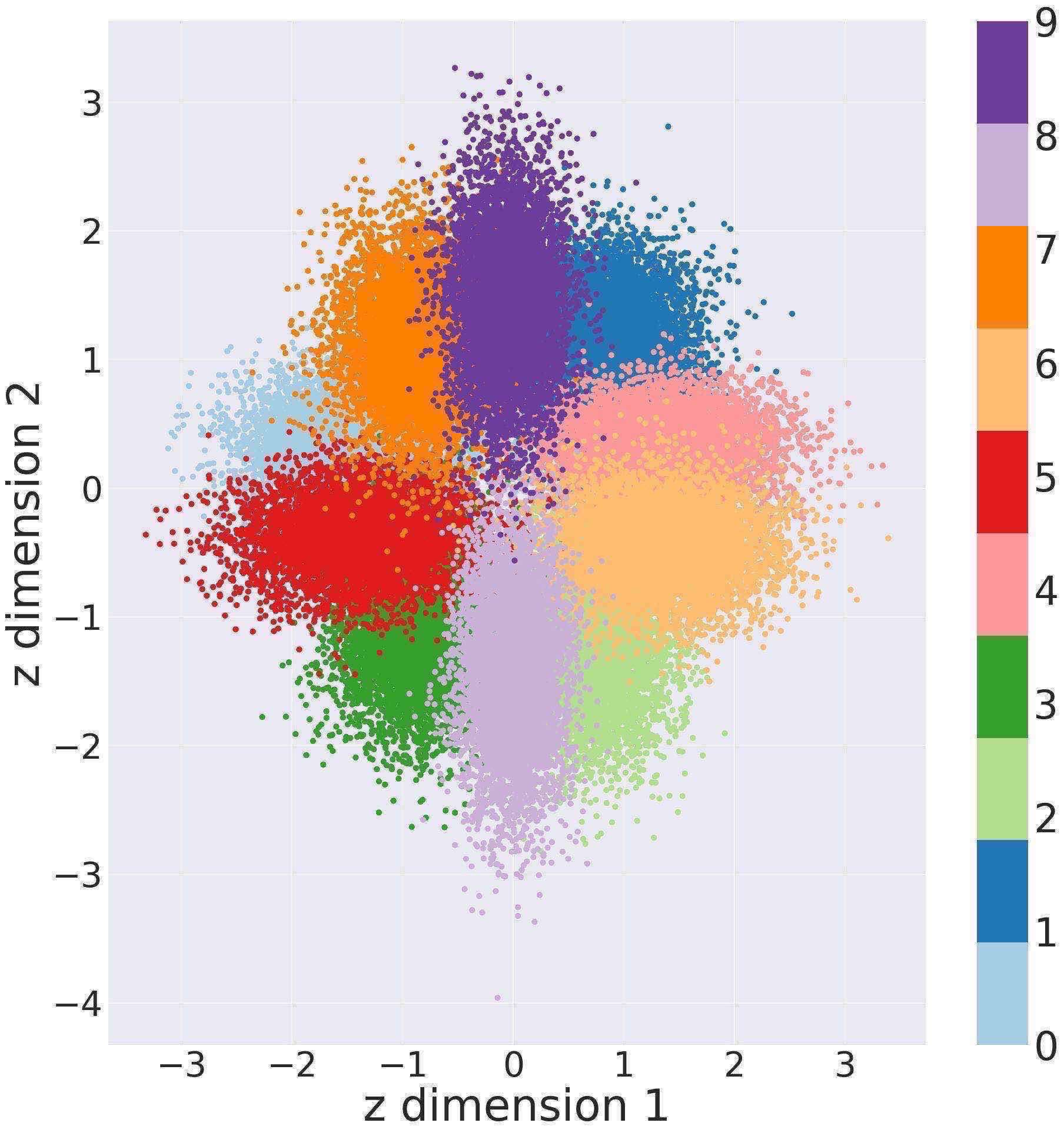}
\quad
\includegraphics[width = 0.22 \columnwidth]{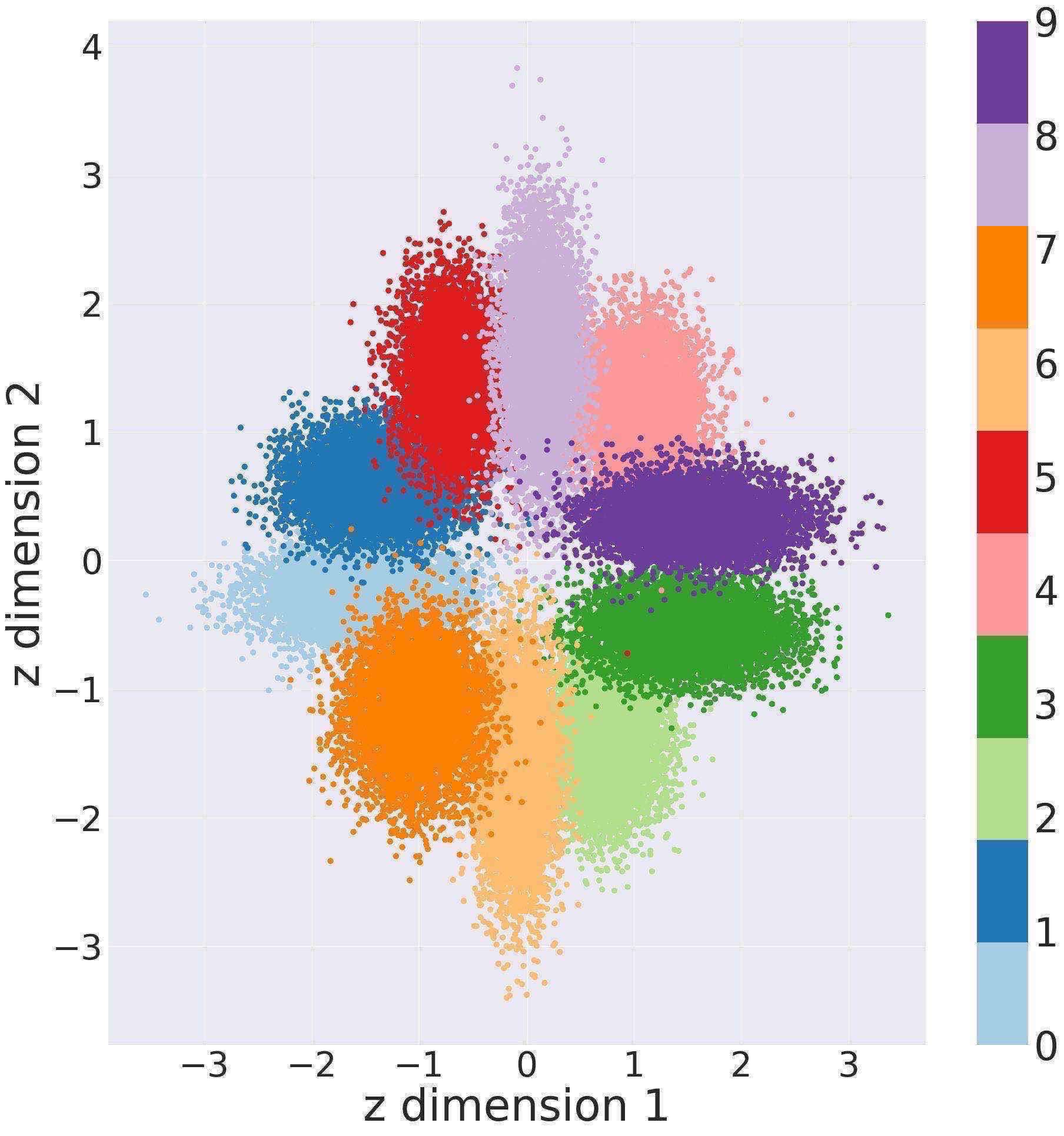}
\quad
\includegraphics[width = 0.22 \columnwidth]{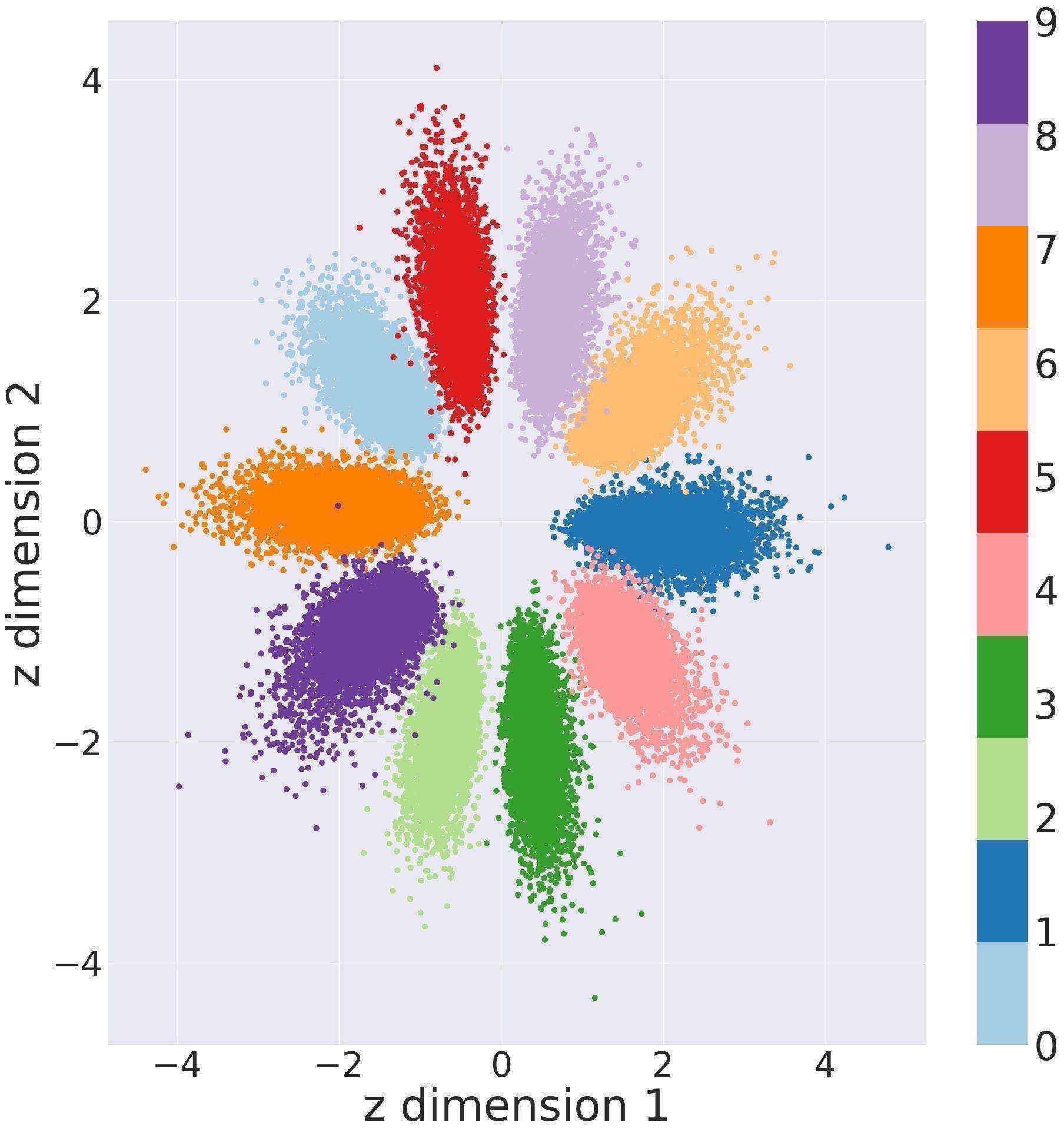}
\quad
\includegraphics[width = 0.22 \columnwidth]{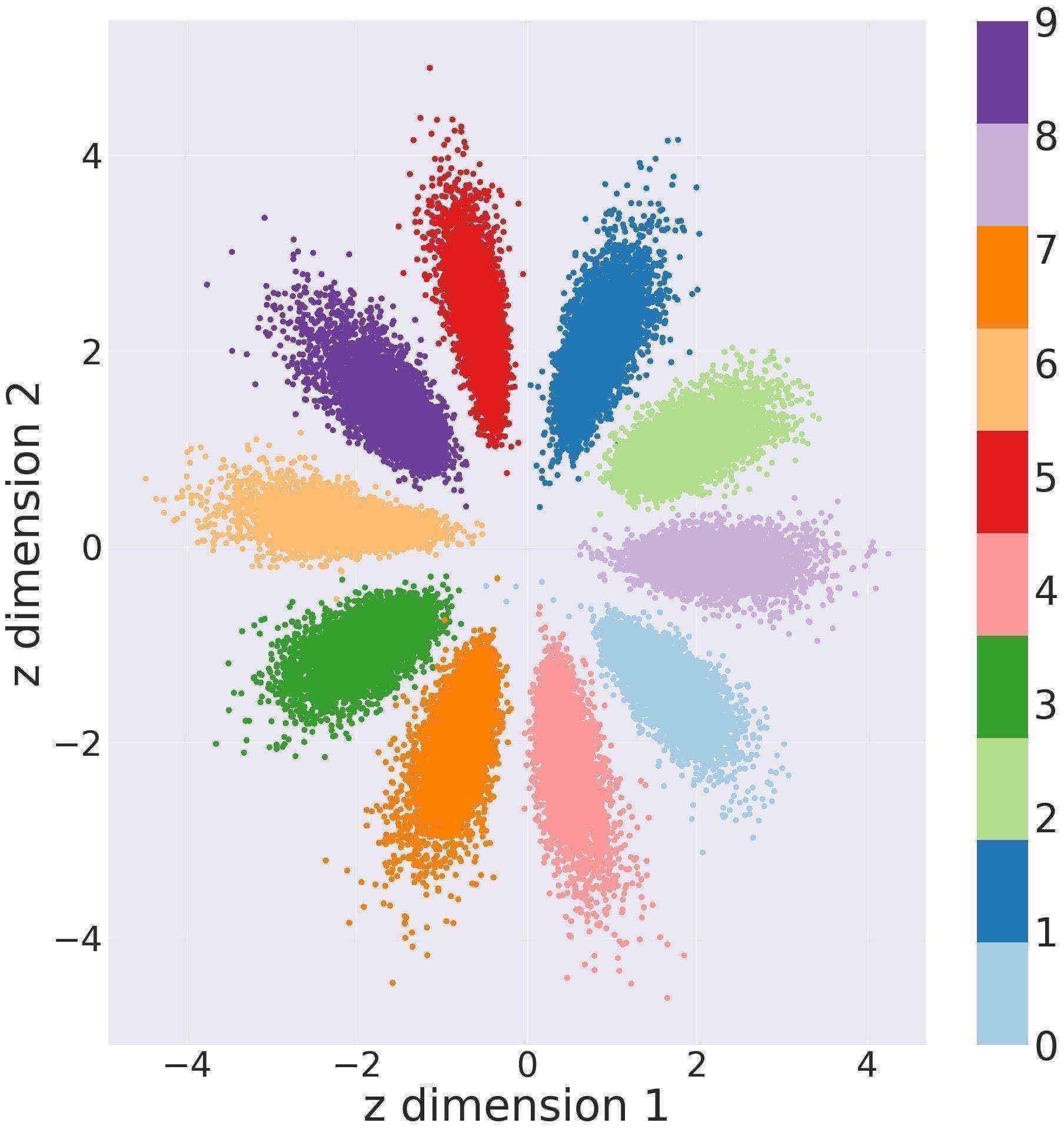}
\caption{2-D MNIST latent space visualization with different $\beta$ values. From left to right, visualization of $\beta = 1.0, 0.5, 0.1, 0.05$.}\label{fig:latent_beta_2D}
\end{figure}

To visualize the effect of beta, we trained multiple models with different $\beta$ values on the MNIST dataset, in an isolated fashion with all data present at all times to focus on the effect of $\beta$. The corresponding two-dimensional aggregate encodings at the end of training are shown in Figure \ref{fig:latent_beta_2D}. Here, we can empirically observe above described phenomenon. With a beta of one and larger, the aggregate posterior's structure starts to collapse and the aggregate encoding converges to a normal distribution.

While this minimizes the distributional mismatch with respect to the prior, the separability of classes is also lost and an accurate classification cannot be achieved. On the other hand, if the beta value becomes ever smaller, there is insufficient regularization present, and the aggregate posterior no longer follows a normal distribution at all. The latter does not only render sampling for generative replay difficult, it also challenges the assumption of distances to each class' latent mean being Weibull-distributed, as the latter can essentially be seen as a skewed normal.

At this point, it is important to make the following note. Whereas the interpretation of beta always follows the  mentioned reasoning, the precise quantitative values of beta can depend heavily on the way losses are treated in practice.~In particular, we emphasize that many coding environments, such as PyTorch and TensorFlow {(\url{https://pytorch.org} and \url{https://www.tensorflow.org}, accessed on 22 January 2022)}, tend to average losses by default, e.g., normalization of the reconstruction loss term by spatial image dimensionality $I \times I$ and respective normalization of the KLD by the model's latent dimensionality.

Arguably, the former factor leads to a much larger division than the latter. As such, the natural scale on which the individual reconstruction and KLD losses operate, with the KLD usually being a much much smaller regularization term, can easily be altered. We thus emphasize that the quantitative value of beta should always be regarded in its exact empirical context.
In order to provide a quantitative intuition for the role of loss normalization and its connection to beta, we show examples for the models trained with different $\beta$ with 2-D latent spaces and 60-D latent spaces in Tables \ref{tab:beta_2d} and \ref{tab:beta_60d}, respectively. In both cases, the losses were normalized by the respective spatial image size and chosen latent dimension, i.e., nats per dimension.

For reference, the un-normalized nats quantities are reported in brackets. We observe that decreasing the value of beta below one is necessary to improve the classification accuracy when the losses are normalized, as well as the overall variational lower bound. Taking the 60 dimensional case as a specific example, we can also observe that reducing the beta value too far and decreasing it from, e.g., $0.1$ to $0.05$ leads to deterioration of the variational lower bound, from $119.596$ to $121.101$ natural units, while the classification accuracy by itself does not improve further.

We can see that this is due to the \emph{KL} divergence residing on the same scale as the normalized reconstruction loss, whereas the latter would typically be much greater when scaled by the image size. Although this may initially appear to render the interpretation of beta more complicated than advocated in the initial work \cite{Higgins2017}, we noticed that a normalized loss appears to come at the advantage of the same beta value of $0.1$ consistently yielding the best results across all of our experiments. That is, we can use the same value of beta independently of whether the $28 \times 28$ MNIST or the $256 \times 256$ flower images are investigated, always with a latent dimensionality of $60$.

\begin{table}[H]

\caption{Losses obtained for different $\beta$ values for MNIST with a 2-D latent space. Training conducted in isolated fashion to quantitatively showcase the role of $\beta$. Un-normalized values in nats are reported in brackets for reference purposes.}
\label{tab:beta_2d}\setlength{\tabcolsep}{4.05mm}
{\small\begin{tabular}{ccccccc}\toprule

& & \multicolumn{3}{c}{\textbf{In Nats per Dimension (Nats in Brackets)}} & \\
\midrule
\textbf{2-D Latent} & \textbf{Beta} & \textbf{KLD} & \textbf{Recon Loss} & \textbf{Class Loss} & \textbf{Accuracy [\%]} \\
\midrule

train & 1.0 & 1.039 (2.078) & 0.237 (185.8) & 0.539 (5.39) & 79.87  \\
test &  & 1.030 (2.060) & 0.235 (184.3) & 0.596 (5.96) & 78.30   \\
\midrule
train & 0.5 & 1.406 (2.812) & 0.230 (180.4) & 0.221 (2.21) & 93.88  \\
test & & 1.382 (2.764) & 0.228 (178.8) & 0.305 (3.05) & 92.07  \\
\midrule
train & 0.1 & 2.055 (4.110) & 0.214 (167.8) & 0.042 (0.42) & 99.68  \\
test & & 2.071 (4.142) & 0.212 (166.3) & 0.116 (1.16) & 98.73 \\
\midrule
train & 0.05 & 2.395 (4.790) & 0.208 (163.1) & 0.025 (0.25) & 99.83  \\
test & & 2.382 (4.764) & 0.206 (161.6) & 0.159 (1.59) & 98.79  \\ \bottomrule

\end{tabular}}
\end{table}
\unskip

\begin{table}[H]

\caption{Losses obtained for different $\beta$ values for MNIST with a 60-D latent space. Training conducted in isolated fashion to quantitatively showcase the role of $\beta$. Un-normalized values in nats are reported in brackets for reference purposes.}
\label{tab:beta_60d}\setlength{\tabcolsep}{3.65mm}
{\small\begin{tabular}{ccccccc}\toprule

& & \multicolumn{3}{c}{\textbf{In Nats per Dimension (Nats in Brackets)}} & \\
\midrule
\textbf{60-D Latent} & \textbf{Beta} & \textbf{KLD} & \textbf{Recon Loss} & \textbf{Class Loss} & \textbf{Accuracy [\%]} \\
\midrule

train & 1.0 & 0.108 (6.480) & 0.184 (144.3) & 0.0110 (0.110) & 99.71\\
test &  & 0.110 (6.600) & 0.181 (142.0) & 0.0457 (0.457) & 99.03 \\
\midrule
train & 0.5 & 0.151 (9.060) & 0.162 (127.1) & 0.0052 (0.052) & 99.87\\
test & & 0.156 (9.360) & 0.159 (124.7) & 0.0451 (0.451) & 99.14 \\
\midrule
train & 0.1 & 0.346 (20.76) & 0.124 (97.22) & 0.0022 (0.022) & 99.95 \\
test & & 0.342 (20.52) & 0.126 (98.79) & 0.0286 (0.286) & 99.38\\
\midrule
train & 0.05 & 0.476 (28.56) & 0.115 (90.16) & 0.0018 (0.018) & 99.95 \\
test & & 0.471 (28.26) & 0.118 (92.53) & 0.0311 (0.311) & 99.34 \\
\bottomrule
\end{tabular} }
\end{table}

\subsection[\appendixname~\thesubsection]{Complementary Generative Modelling Advances}\label{app:sec_gen_advances}
At the time of their initial introduction, it was notorious that variational autoencoders produced blurry examples and were associated with an inability to scale to more complex high-resolution color images. This is in contrast to their prominent generative counterparts, the generative adversarial network \cite{Goodfellow2014}. Although this stigma perhaps still holds until today, there have been many successful recent efforts to address this challenge.

In our final outlook in the main body, we thus empirically showcased the impact of generative modeling advances with their optional improvements in two promising research directions: autoregression \cite{Oord2016, Gulrajani2017, Chen2016} and introspection \cite{Huang2018, Ulyanov2018}. The commonality between these approaches is their aim to overcome the limitations of independent pixel-wise reconstructions. In this appendix section, we  briefly summarize the foundation of these generative~extensions.

\subsubsection{Improvements through Autoregressive Decoding}
In essence, autoregressive models improve the probabilistic decoder through a spatial conditioning of each scalar output value on the previous ones, in addition to conditioning on the latent variable:
\begin{equation}\label{eq:autoregressive}
p(\boldsymbol{x} | \boldsymbol{z}) = \prod_{i} p(x_{i} | x_{1}, \ldots, x_{i - 1}, \boldsymbol{z})
\end{equation}

In an image, generation thus needs to proceed pixel by pixel and is commonly referred to as PixelVAE \cite{Gulrajani2017}. This conditioning is generally achieved by providing the input to the decoder during training, i.e., including a skip path that bypasses the probabilistic encoding. A concurrent introduction of autoregressive VAEs  thus coined this model ``lossy'' \cite{Chen2016}. This is because local information can now be increasingly modeled without access to the latent variable, and the encoding of $\boldsymbol{z}$ can focus on the global information.

Although the main body's accuracies of generative replay with autoregression are assuring in the MNIST scenario, autoregressive sampling comes with a major caveat. When attempting to operate on larger data, the computational complexity of the pixel by pixel data creation procedure grows in direct proportion to the input dimensionality. With increasing input size, the repeated calculation of the autoregressive decoder layers can thus rapidly render the generation practically infeasible.

\subsubsection{Introspection and Adversarial Training}
A promising alternative perspective towards autoencoding beyond pixel similarities is to leverage the insights obtained from generative adversarial networks (GAN). To this matter, Larsen et al. \cite{Larsen2016}  proposed a hybrid model called VAE-GAN. Here, the crucial idea is to append a GAN style adversarial discriminator to the variational autoencoder. This yields a model that promises to overcome a conventional GAN's mode collapse issues, as the VAE is responsible for the rich encoding, while letting the added discriminator judge the decoder's output based on perceptual criteria rather than individual pixel values.

The more recent IntroVAE \cite{Huang2018} and adversarial encoder generator networks \cite{Ulyanov2018} have subsequently come to the realization that this does not necessarily require the auxiliary real-fake discriminator, as the VAE itself already provides strong means for discrimination, namely its probabilistic encoder. We leverage this idea of introspection for our framework, as it does not require any architectural or structural changes beyond an additional term in the loss function.

For sake of brevity, we denote the probabilistic encoder through the parameters $\boldsymbol{\phi}$ and decoder $\boldsymbol{\theta}$ in the following equations. Training our model with introspection is then equivalent to adding the following two terms to our previously formulated loss function:

\begin{equation} \label{eq:introvae_loss_encoder}
\mathcal{L}_{IntroVAE\_Enc} = \mathcal{L}_{VAE} - \beta \left[ m - \kld{\boldsymbol{\theta}(\boldsymbol{\phi}(\boldsymbol{z}))}{p(\boldsymbol{z})} \right]^{+}
\end{equation}
and
\begin{equation} \label{eq:introvae_loss_decoder}
\mathcal{L}_{IntroVAE\_Dec} =  \mathcal{L}_{Rec} - \beta \kld{\boldsymbol{\theta}(\boldsymbol{\phi}(\boldsymbol{z}))}{p(\boldsymbol{z})} \, .
\end{equation}

Here, $\mathcal{L}_{VAE} $ corresponds to the full loss of the main body's Equation (1) and $\mathcal{L}_{Rec}$ corresponds to the reconstruction loss portion: $\mathbb{E}_{q_{\boldsymbol{\theta}}(\boldsymbol{z} | \boldsymbol{x}^{(n)})} [ \log{p_{\boldsymbol{\phi}}(\boldsymbol{x}^{(n)} | \boldsymbol{z})} ] $. In the above equations, we followed the original authors' proposal to include a positive margin $m$, with $[\cdot ] $ denoting $max(0, \cdot)$. This hinge loss formulation serves the purpose of empirically limiting the encoder's reward to avoid a too massive gap in a min--max game of above competing \emph{KL} terms.

Aside from the regular loss that encourages the encoder to match the approximate posterior to the prior for real data, the encoder is now further driven to maximize the deviation from the posterior to the prior for generated images. Conversely, the decoder is encouraged to ``fool'' the encoder into producing a posterior distribution that matches the prior for these generated images. The optimization is conducted jointly. In comparison with a traditional VAE, this can thus be seen as training in an adversarial-like manner, without necessitating additional discriminative models. As such, introspection fits naturally into our OpenVAE, and no further changes are required.

\subsection[\appendixname~\thesubsection]{Training Hyper-Parameters and Architecture Definitions}\label{app:sec_hyperparams}

In this section, we provide a full specification of hyper-parameters, model architectures and the training procedure used in the main body.

\subsubsection*{Architecture} For our MNIST style continual-learning experiments, we report both a simple multi-layer perceptron architecture, as well as a deeper wide residual network variant. For the former, we follow previous continual-learning studies and employ a two-hidden-layer and 400-unit multi-layer perceptron \cite{Kemker2018}. For the latter, we base our encoder and decoder architecture on 14-layer wide residual networks \cite{He2016, Zagoruyko2016} with a latent dimensionality of $60$ to demonstrate scalability to high-dimensions and as used in lossy auto-encoders \cite{Gulrajani2017, Chen2016}. Our main body's reported out-of-distribution detection experiments are all based on this WRN architecture.

For a common frame of reference, all methods share the same underlying WRN architecture, including the investigated separate classifiers (for OpenMax) and generative models of the reported dual model approaches. All hidden layers include batch-normalization \cite{Ioffe2015} with a value of $10^{-5}$ and use rectified linear unit (ReLU) activations. A detailed list of the architectural components is provided in Tables \ref{tab:WRNEnc_ArcDefinition} and \ref{tab:WRNDec_ArcDefinition}. For the higher resolution $256 \times 256$ flower images, we used a deeper 26-layer WRN version, in analogy to previous works \cite{Gulrajani2017, Chen2016}. Here, the last encoder and first decoder blocks are repeated an extra three times, resulting in an additional three stages of down- and up-sampling by a factor of two. The encoder's spatial output dimensionality is thus equivalent to the 14-layer architecture applied to the eight-times lower resolution images of the simpler datasets.

\subsubsection*{Autoregression} For the autoregressive variant, we set the number of output channels of the decoder to 60 and append three additional pixel decoder layers, each with a kernel size of $7 \times 7$. We use 60 channels in each autoregressive layer for the MNIST dataset and 256 for the more complex flower data \cite{Gulrajani2017, Chen2016}. Whereas we report reconstruction log-likelihoods in natural units (nats) in the upcoming detailed supplementary results (recall that we have only shown quantitative validation of the model through the proxy of continual classification in the main body), these models are practically formulated as a classification problem with a 256-way Softmax. The corresponding loss is in bits per dimension.

We converted these values to have a better comparison; however, in order to do so, we need to sample from the pixel decoder's multinomial distribution to calculate a binary cross-entropy on reconstructed images. We further note that all losses are normalized with respect to the spatial and latent dimensions, as mentioned in the prior appendix section, which  explained normalization in the context of the role of beta.

\begin{table}[H]
\caption{A 14-layer wide residual network (WRN) encoder with a widen factor of 10. Convolutional layers (conv) are parametrized by a quadratic filter size followed by the amount of filters. p
and s represent zero padding and stride, respectively. If no padding or stride is
specified, then p = 0 and s = 1. Skip connections are an additional operation at a layer, with the layer to be skipped specified in brackets. Convolutional layers are followed by batch-normalization and a rectified linear unit (ReLU) activation. The probabilistic encoder ends on fully-connected layers for $\mu$ and $\sigma$ that depend on the chosen latent space dimensionality and the data's spatial size. }
\label{tab:WRNEnc_ArcDefinition}
\setlength{\tabcolsep}{4.8mm}
{\small\begin{tabular}{ll}
\toprule
\textbf{Layer Type} & \textbf{WRN Encoder}\\
\midrule

Layer 1 & conv $3\times3$---48, p = 1\\  \midrule
Block 1 & \begin{tabular}{@{}l@{}l@{}}conv $3\times3$---160, p = 1;&\quad\quad\quad\quad\hspace{1.5em}  conv $1\times1$---160 (skip next layer)\\ conv $3\times3$---160, p = 1&\\ conv $3\times3$---160, p = 1;&\quad\quad\quad\quad\hspace{1.5em} shortcut (skip next layer)\\ conv $3\times3$---160, p = 1&\end{tabular}\\
\midrule
Block 2 & \begin{tabular}{@{}ll@{}} conv $3\times3$---320, s = 2, p = 1;&  conv $1\times1$---320, s = 2 (skip next layer)\\ conv $3\times3$---320, p = 1&\\ conv $3\times3$---320, p = 1;& shortcut (skip next layer)\\ conv $3\times3$---320, p = 1&\end{tabular}\\
\midrule
Block 3 & \begin{tabular}{@{}ll@{}} conv $3\times3$---640, s = 2, p = 1;& conv $1\times1$---640, s = 2 (skip next layer)\\ conv $3\times3$---640, p = 1&\\ conv $3\times3$---640, p = 1;& shortcut (skip next layer)\\ conv $3\times3$---640, p = 1&\end{tabular}\\
\bottomrule
\end{tabular}}
\end{table}
\unskip
\begin{table}[H]
\caption{A 14-layer WRN decoder with a widen factor of 10. $P_w$ and $P_h$ refer to the input's spatial dimension. Convolutional (conv) and transposed convolutional (conv\_t) layers are parametrized by a quadratic filter size followed by the amount of filters. p and s represent zero padding and stride, respectively. If no padding or stride is
specified, then p = 0 and s = 1. Skip connections are an additional operation at a layer, with the layer to be skipped specified in brackets. Every convolutional and fully-connected (FC) layer is followed by batch-normalization and a rectified linear unit (ReLU) activation function. The model ends on a Sigmoid function.}
\label{tab:WRNDec_ArcDefinition}\setlength{\tabcolsep}{6.2mm}
{\small\begin{tabular}{ll}
\toprule
\textbf{Layer Type} & \textbf{WRN Decoder}\\
\midrule

Layer 1 & FC $640\times\floor*{P_w/4}\times\floor*{P_h/4}$\\
\midrule
Block 1 & \begin{tabular}{@{}l@{}l@{}}conv\_t $3\times3$---320, p = 1;&\quad\quad\quad\quad\hspace{0.5em}  conv\_t $1\times1$---320 (skip next layer)\\ conv $3\times3$---320, p = 1&\\ conv $3\times3$--- 320, p = 1;&\quad\quad\quad\quad\hspace{0.5em} shortcut (skip next layer)\\ conv $3\times3$---320, p = 1&\\
upsample $\times$ 2\end{tabular}\\
\midrule
Block 2 & \begin{tabular}{@{}l@{}l@{}}conv\_t $3\times3$---160, p = 1;&\quad\quad\quad\quad\hspace{0.5em}  conv\_t $1\times1$---160 (skip next layer)\\ conv $3\times3$---160, p = 1&\\ conv $3\times3$--- 160, p = 1;&\quad\quad\quad\quad\hspace{0.5em} shortcut (skip next layer)\\ conv $3\times3$---160, p = 1&\\
upsample $\times$ 2\end{tabular}\\
\midrule
Block 3 & \begin{tabular}{@{}l@{}l@{}}conv\_t $3\times3$---48, p = 1;&\quad\quad\quad\quad\hspace{1em}  conv\_t $1\times1$---48 (skip next layer)\\ conv $3\times3$---48, p = 1&\\ conv $3\times3$---48, p = 1;&\quad\quad\quad\quad\hspace{1em} shortcut (skip next layer)\\ conv $3\times3$---48, p = 1&\end{tabular}\\
\midrule
Layer 2 & conv $3\times3$---3, p = 1\\
\bottomrule
\end{tabular} }
\end{table}

\subsubsection*{Introspection} For the introspective model variant, we note that the original authors of IntroVAE  introduced additional weighting terms in front of the reconstruction loss, in order to drastically lower its magnitude and the added \emph{KL} divergence. We  observed that the former is simply due to lack of normalization with respect to input width and height and hence the reconstruction loss growing proportionally with the spatial input size, whereas the \emph{KL} divergence typically does not reflect this behavior for a fixed-size latent space.

Given that we average the loss over the image size in our practical experimentation, we found this additional hyper-parameter to be unnecessary. The other hyper-parameter to weight the added adversarial \emph{KL} divergence term is essentially equivalent to the already introduced beta, alas without our motivation in earlier sections but simply as a heuristic so as to not overpower the reconstruction loss. The introspective variant of our OpenVAE thus does not introduce any additional hyper-parameters or architectural modifications beyond the additional loss term, as summarized in the previous section.

\subsubsection*{Stochastic Gradient Descent (SGD)} Optimization parameters were used in consistence with the literature \cite{Gulrajani2017, Chen2016}. Accordingly, all models are optimized using stochastic gradient descent with a mini-batch size of $128$ and Adam \cite{Kingma2015} with a learning rate of $0.001$ and first and second momenta equal to $0.9$ and $0.999$. For MNIST, FashionMNIST and AudioMNIST, no data augmentation or preprocessing is applied. For the flower experiments, images are stochastically flipped horizontally with a 50\% chance and the mini-batch size is reduced to $32$. We initialize all weights according to He et al. \cite{He2015}. All class incremental models were trained for 120~epochs per task, except for the flower experiment. While our investigated single model exhibits representational transfer due to weight sharing and need not necessarily be trained for the entire amount of epochs for each subsequent task, this guarantees convergence and a fair comparison of results with respect to the achievable accuracy of other methods.

Due to the much smaller dataset size, architectures were trained for $2000$ epochs on the flower images, in order to obtain a similar amount of update iterations. For the generative replay with statistical outlier rejection, we used an aggressive rejection rate of $\Omega_{t} = 0.01$ (although we obtain almost analogous results with $0.05$) and dynamically set tail-sizes to 5\% of seen examples per class. As mentioned in the main body, the used open-set distance measure was the cosine distance.

To enable our out-of-distribution detection experiments in the main body---which includes comparison with datasets, such as CIFAR---we resized all images to $32 \times 32$. Recall that we also made use of the AudioMNIST dataset in the main body to showcase the challenge in open-set recognition, where most approaches fail to recognize Audio data as out-of-distribution, even though its form is entirely different from the commonly observed object-centric images. To make this comparison possible, we followed the original dataset's authors and use the described Fourier transform on the audio data to obtain frequency~images.

\subsubsection*{De-Quantization, Overfitting and Data Augmentation} As the autoregressive model variants require a de-quantization of the input (to transform the discrete eight-unit input into a continuous distribution) \cite{Gulrajani2017, Chen2016}, we employed a denoising procedure on the input. Specifically, for our MNIST-like gray-scale datasets, we add noise to the input, sampled from a normal distribution with mean $0$ and variance $0.25$. As typical for denoising autoencoders, the reconstruction loss nevertheless aims to recover the unperturbed original input.

As the latter could be argued to serve an additional data augmentation effect (we always observed an improvement), we adopted the denoising procedure for all models, even if no autoregression was used.
In this way, a fair comparison was enabled. For the colored high-resolution flower images, such gray-scale Gaussian noise seems less meaningful. Here, we realize that our primarily interest lies in maintaining the discriminative performance of our model and less so on the visual quality of the generated data.

We can thus take advantage of the de-quantization perturbation distribution as means to encode our prior knowledge of common generative pitfalls.
In our specific context, it is well known that a traditional VAE without further advances commonly fails to generate non-blurry, crisp images. However, we can include and work around this belief by letting the denoising assume the form of de-blurring, e.g., by stochastically adding a varying Gaussian blur to inputs (as done in \cite{Hendrycks2019}).

Even though the decoder is ultimately still encouraged to remove this blur and reconstruct the original clean image, the encoder is now inherently required to learn how to manage blurry input. It is encouraged to build up a natural invariance to our choice of perturbation. In the context of maintaining a classifier with generative replay, to an extent, it should then no longer be a strict requirement to replay locally detailed crisp images, as long as the information required for discrimination is present.

\subsubsection*{Hardware and Software} All models were trained on single GeForce GTX 1080 (Nvidia, Santa Clara, CA, USA) graphics processing units (GPU), with the exception of the high-resolution flower image experiments, where we used a single V100 GPU (Nvidia, Santa Clara, CA, USA) per experiment. \xadded{Our implementation is based on \textit{PyTorch} {(\url{https://pytorch.org}, last accessed on 22 January 2022}),
including data loading functionality for the majority of the investigated public datasets through the \textit{torchvision} library. The AudioMNIST data was preprocessed using the \textit{librosa} Python library (\url{https://librosa.org/doc/latest/index.html}, last accessed on 22 January 2022), following the setting of the original AudioMNIST dataset authors~\cite{Becker2018}. Our code will be publicly available.}

\subsubsection*{Elastic Weight Consolidation (EWC)} Recall that for related work approaches, we reported quantitative values found in the literature if our reproduction matched or did not surpass this number or, conversely, our obtained value if it turned out to be better. Primarily, the latter discrepancy was the case for our reproduction of the EWC experiments on FashionMNIST, where we obtained marginally improved results. Here, the number of Fisher samples was fixed to the total number of data points from all the previously seen tasks. A suitable Fisher multiplier value $\lambda$ was determined by conducting a grid search over a set of five values: 50, 100, 500, 1000 and 5000 on held-out validation data for the first two tasks in sequence. We observed exploding gradients if $\lambda$ was too high. However, a very small $\lambda$ lead to excessive drift in the weight distribution across subsequent tasks that further resulted in catastrophic interference. Empirically, $\lambda = 500$ seemed to provide the best balance.

\subsection[\appendixname~\thesubsection]{Limitations}\label{app:sec_limitations}
We believe that there are three main sources for limitations of our work. Some of these are general to the considered continual-learning scenarios, whereas others are more directly related to our proposed approach. Specifically, we can group the limitations into:

\begin{enumerate}
\item Limitations of our proposed aggregate posterior-based EVT approach and its use for open-set recognition and generative replay.
\item Limitations of the employed generative model variant, i.e., caveats of autoregression or introspection.
\item Limitations in terms of obtainable insights from investigated scenarios, i.e.,~the specific continual-learning set-up.
\end{enumerate}

As only the first point is concerned with immediate limitations of our method, we provide the most detailed discussion of these aspects first, before providing a small overview of conceivable general limitations that are less specific to the contributions of our paper but, nevertheless, of potential interest.

\subsubsection{Aggregate Posterior-Based EVT Limitations}
There are several imaginable caveats to our method, some of which are of theoretical nature and have not yet been observed in our experiments and some of which the reader should be aware of when reproducing our experiments. The two main caveats we surmise are: the assumed unimodality of the latent space distance distribution that forms the basis for the Weibull based EVT approach and the necessity for a ``burn-in'' phase at the start of~training.

\textit{Distance distribution unimodality:} Recall that we make use of extreme value theory by fitting a Weibull distribution based on the distances to the practically obtained aggregate posterior of our joint model. This was motivated from a direct mathematical expression for the aggregate posterior being cumbersome to obtain, as the latter can in principle be arbitrarily complex. To circumvent this challenge, we imposed a linear separation of classes through the use of a linear classifier on the latent variable $\boldsymbol{z}$ and subsequent treatment of the aggregate posterior-based on each individual class. As such, we obtained the distances to the mean of the aggregated encoding for each class. As a result, a Weibull distribution for statistical outlier rejection was crafted, with one mode per class. While this Weibull distribution can be multivariate depending on the choice of distance distribution, e.g., a cosine distance would collapse latent vectors into scalar distances and a euclidean distance could preserve dimensionality, each class is nevertheless assumed to have a single distance mean and thus a single mode of the Weibull distribution. This forms a  theoretical limitation of our approach, although not presently observed to hinder our practical application.

Concretely, we assume that the distances to the mean can be described by a single mean of the distribution, a single variance and a shift parameter. Intuitively, this limits the applicability of the approach, should the obtained aggregate posterior per class remain more complex in terms of forming multiple clusters within a class. However, we also note that a unimodality in terms of distance distribution is not analogous to the existence of a single well-formed cluster in the aggregate posterior. For instance, if a class were to be symmetrically distributed around a low-density region, think of a donut for example, the three parameter distribution on the distances to the center would still capture this through a single mode. A limitation would arise if multiple clusters within a class were to form without the presence of such symmetry. We have not yet observed the latter in practice; however, we note that it presently marks a theoretical limitation of our specified approach.

\textit{``Burn-in'' phase:} Our obtained Weibull distance distribution based on the aggregate posterior is obtained as a ``meta-recognition'' module. That is, the Weibull distribution is not trained but is a derived quantity of the aggregate posterior. In order to form the basis for statistical rejection of outliers and constraining generation to inliers an initial ``burn-in'' phase needs to exist to first obtain some meaningful approximate posterior estimate. At this point, it could be argued whether this is a limitation or not. In principle, any deep neural network model arguably has to undergo an initial stage of training before its representations can be leveraged. As one of the main contributions in our paper is improved open-set recognition, we believe this aspect is nevertheless important to mention. Consequently, our trained model and aggregate posterior-based open-set-recognition mechanism enable robust application of a trained model or, as demonstrated, its continual learning. In the very first epochs of training, it is however presently expected that the data reflects the true task and does not contain potentially corrupted inputs, as a notion of ``in-distribution'' first needs to be built up. Even though we mention this as a limitation, we also note that we are unaware of a deep-neural-network-based approach that would not have an initial training phase as a necessity in supervised learning.

\subsubsection{Limitations of the Employed Generative Model Variants}
We investigated modern generative modeling advances that build on top of the conventional VAE. The primary purpose was to demonstrate that the notorious blurriness of VAE generations can easily be overcome based on recent insights. As such, our proposed approach was shown in the context of more complex high-resolution color images, with recent generative modeling advances being shown to draw similar benefits from our proposed open-set mechanism. Two investigated variants were autoregression (PixelVAE) and introspection (IntroVAE). Although neither of these formulations is our contribution, nor a key to our proposed formulation, we provide a brief description of their limitations in a continual-learning context.

\textit{Autoregression:} During training, autoregression does not initially appear to come with significant caveats beyond the perhaps added computational overhead of using larger sized convolutions to capture more local context (e.g., $7 \times 7$ convolutions in contrast to typically employed stacks of $3 \times 3$ kernels). The conditioning on pixel values during training is usually practically achieved through masking operations, enabling training on a similar time scale to non-autoregressive counterparts. However, for the generation of actual images, pixel values need to be sampled sequentially, much in contrast to a typical VAE calculating the entire decoder in a single-shot pass. As such, the time it takes to continually learn on later tasks, where old information is rehearsed based on generative examples, comes with an unfortunate increase in required computation time taken up for autoregressive sampling. For plain VAEs, or the IntroVAE variant, this is not usually a problem as generation typically takes significantly less time than training with backpropagation. For autoregressive sampling in its sequential formulation, this is quickly no longer the case.

\textit{Introspection:} Fortunately, introspection does not come with a similar computational caveat as autoregression. In contrast to a conventional VAE, the computational overhead lies in one additional pass through the encoder for each update to compute the adversarial term of generated examples. As such, this computational increase is a less severe drawback. A perhaps more significant caveat is the presence of the min--max objective potentially rendering the training more difficult in terms of finding a good point of convergence. In other words, because the losses are balanced in the adversarial game, finding a satisfactory end point can often be subject to subjectively being satisfied with the visual perceived generation quality. Whereas the IntroVAE benefits greatly from stability (in contrast to the often observed collapse in pure GANs) and is thus typically trained for extended periods of time, this could also mark a conceivably difficult trade-off when deciding when to continue optimizing the next task in continual learning while simultaneously minimizing total training time.

\subsubsection{Limitations of the Investigated Scenarios}
The key contribution of our paper is in showing how a principled single mechanism can be used in a single model to unify the prevention of catastrophic interference in continual learning with open-set recognition. The present formulation of our paper investigates these aspects in two experimental subsections, to adequately showcase the benefits from each perspective. In retrospect, while this is the initially stated motivation, it is clear that the investigated challenges are actually part of a greater theme towards robust continual~learning.

As there is no immediate literature to compare with, we decided that the presented empirical analysis of the main body would provide the most immediate benefit to the reader. We do however also note that this is a more general limitation of predominant investigation in the literature. Here, our work provides the first steps in the direction of a more meaningful investigation of continual learning, for instance, where task scenarios are not pre-defined to contain clear-cut boundaries. Our approach has demonstrated that it is possible to accurately identify when the distribution experiences a major disruption in the process of learning continually. Future investigation should thus lift persisting limitations of the rigid investigation and consider scenarios where tasks are not always introduced at a known point in time.

\subsection[\appendixname~\thesubsection]{Full Continual Learning Results for All Intermediate Steps}\label{app:full_results}
In the main body, we reported the final classification accuracy at the end of multiple task increments in continual learning. In this section, we provide a full list of intermediate results and a two-fold extension to the reporting.

First, instead of purely reporting the final obtained accuracy value, we follow prior work \cite{Kemker2018} and report more nuanced accuracy metrics: the base task's accuracy over time $\alpha_{t, base}$, the new task's accuracy $\alpha_{t, new}$ and the overall accuracy at any point in time $\alpha_{t, all}$. The first of these, the ``base'' metric, reports the accuracy of the initial first tasks, e.g., digits 0 and 1 in MNIST and their accuracy degradation over time when the data are no longer available when subsequent tasks arrive. Conversely, the ``new'' metric always portrays only the accuracy of the most recent task, in independence of the other existing classes.

Finally, the ``all'' accuracy is the accuracy averaged over all presently existing tasks. For instance, the final accuracy reported in the main body is thus this overall metric at the end of observing all tasks. This is a more appropriate way to evaluate the quality of the model over time. Given that the employed mechanism to avoid catastrophic interference in continual learning is generative replay, it also gives us further insight into whether an accuracy degradation is due to old tasks being forgotten, i.e., catastrophic interference occurring because the decoder-sampled data will no longer resemble the instances of the observed data distribution or the encoder not being able to encode further new knowledge.

Second, we report respective three metric variants for all intermediate steps of our models for the data negative log-likelihood (NLL). Here, we note that it is particularly important to see the new, base and all metrics in conjunction, as each individual task does not have the same level of difficulty and measures up differently in terms of quantitative NLL values. Nevertheless, the initial task's degradation can similarly be monitored and the overall value at any point in time gives us a direct mean for comparison across models.

In addition to the values reported in the main body, we report the detailed full set of intermediate results for the five task steps of the class incremental scenarios in \mbox{Tables \ref{tab:incremental_MNIST_all}--\ref{tab:incremental_AudioMNIST_all}}. The upper bound (UB) and fine tuning (FT) (tuning on only the most recent task) are again reported for reference. We have now also included a variant of DGR, i.e., a dual model approach with separate generative and separate discriminative model, where the generative model is based on the autoregressive PixelVAE. We omitted the latter result from the main body as the insight is analogous to that of the non-autoregressive comparison (and  comparison to a GAN as the generative model).

The joint open-set model variant appears to have the upper hand, in addition to being able to solve the open-set recognition task. In general, once more, we can observe the increased effect of error accumulation due to unconstrained generative sampling from the prior in comparison to the open-set counterpart that limits sampling to the aggregate posterior. The statistical deviations across experiment repetitions in the base and the overall classification accuracies are higher and are generally decreased by the open-set models. For example, in \mbox{Table \ref{tab:incremental_MNIST_all}} the MNIST base and overall accuracy deviations of a naive supervised variational auto-encoder (SupVAE) are higher than the respective values for OpenVAE, starting already from the second task increment.

Correspondingly, the accuracy values themselves experience larger decline for SupVAE than for OpenVAE with progressive increments. This difference is not as pronounced at the end of the first task increment because the models have not been trained on any of their own generated data yet. Successful literature approaches, such as the variational generative replay proposed by \cite{Farquhar2018}, thus avoid repeated learning based on previous generated examples and simply store and retain a separate generative model for each task.

The strength of our model is that, instead of storing a trained model for each task increment, we are able to continually keep training our joint model with data generated for all previously seen tasks by filtering out ambiguous samples from low-density areas of the posterior. Similar trends can also be observed for the respective pixel models.
Interestingly, the audio dataset also appears to be a prime example to advocate the necessity of a single joint model, rather than maintenance of multiple models as proposed in DGR \cite{Shin2017} or VGR \cite{Farquhar2018}. If we look carefully at the averaged ``all'' accuracy values of our OpenVAE model, we can see that the accuracy between tasks 2 and 3 and, similarly, tasks 4 and 5, first decreases and then increases again. In other words, due to the shared nature of the representations, learning later tasks brings benefits to the solution of already learned former tasks.

Such a form of ``backward transfer'' is difficult to obtain, if not even impossible, in approaches that maintain multiple separated models or even regularization approaches that discourage retrospective change of older representations. We believe the possibility to retrospectively improve older representations to be an additional strength of our approach, where the benefits of a shared representation single model of generative nature become even more evident.

With respect to the obtained negative log-likelihoods we can make two observations. First, by themselves, the small relative improvements between models should be interpreted with caution as they do not directly translate to maintaining continual-learning accuracy. Second, we can also observe that, at every increment for all $\gamma_{t, all}$ and respective quantities for only the new task $\gamma_{t, new}$, negative log-likelihoods are more difficult to interpret compared with the accuracy counterpart. While the latter is normalized between zero and unity, the NLL of different tasks is expected to fluctuate largely according to the task's images' complexity.

To give a concrete example, it is rather straightforward to come to the conclusion that a model suffers from limited capacity or lack of complexity if a single newly arriving class cannot be classified well. In the case of NLL, it is common to observe either a large decrease for the newly arriving class or a large increase depending on the specifically introduced class. As such, these values are naturally comparable between models but are challenging to interpret across time steps without also analyzing the underlying nature of the introduced class.

The exception is formed by the base task's $\gamma_{t, base}$. In analogy to base classification accuracy, this quantity still measures the amount of catastrophic interference across time. However, in all tables we can observe that catastrophic interference is almost imperceptible. As this is not at all reflected in the respective accuracy over time, it further underlines our previous arguments that NLL is not necessarily the best metric to monitor in the presented continual-learning scenario, with the classification proxy seemingly providing a better indicator of continual  generative model degradation.

\begin{table}[H]
\caption{The results for class incremental continual-learning approaches averaged over five runs, baselines and the reference isolated learning scenario for MNIST at the end of every task increment. This is an extension of Table \ref{tab:incremental_results} in the main body. Here, in addition to the accuracy $\alpha_{t}$, $\gamma_{t}$ also indicates the respective negative log-likelihood (NLL) at the end of every task increment $t$. }
\label{tab:incremental_MNIST_all}
\setlength{\tabcolsep}{3.2mm}
\begin{adjustwidth}{-\extralength}{0cm}
{\small\begin{tabular}{lllllllll}\toprule

\textbf{MNIST} & \textbf{\emph{t}} & \textbf{UB} & \textbf{FT} & \textbf{SupVAE} & \textbf{OpenVAE} & \textbf{PixelVAE DGR} & \textbf{SupPixelVAE} & \textbf{OpenPixelVAE} \\
\midrule

\multirow{4}{*}{$\boldsymbol{\alpha_{base,t}}$} & 1 & 100.0 & 100.0 & $99.97~{\pm~0.029}$ & 99.98 ${\pm~ 0.018}$ & 99.97  ${\pm~ 0.002}$ & 99.97 ${\pm~ 0.026}$ & 99.86 ${\pm~ 0.084}$ \\
& 2 & 99.82 & 00.00 & 97.28 ${\pm~ 3.184}$ & 99.30 ${\pm~ 0.100}$ & 99.54 ${\pm~ 0.285}$ & 96.90 ${\pm~ 2.907}$ & 99.64 ${\pm~ 0.095}$ \\
& 3 & 99.80 & 00.00 & 87.66 ${\pm~ 8.765}$ & 96.69 ${\pm~ 2.173}$ & 99.16 ${\pm~ 0.611}$ & 90.12 ${\pm~ 5.846}$ & 98.88 ${\pm~ 0.491}$\\
(\%) & 4 & 99.85 & 00.00 & 54.70 ${\pm~ 22.84}$ & 94.71 ${\pm~ 1.792}$ & 98.33 ${\pm~ 1.119}$ & 76.84 ${\pm~ 9.095}$ & 98.11 ${\pm~ 0.797}$\\
& 5 & 99.57 & 00.00 & 19.86 ${\pm~ 7.396}$ & 92.53 ${\pm~ 4.485}$  & 98.04 ${\pm~ 1.397}$ & 56.53 ${\pm~ 4.032}$ & 97.44 ${\pm~ 0.785}$\\
\midrule
\multirow{4}{*}{$\boldsymbol{\alpha_{new,t}}$} & 1 & 100.0 & 100.0 & 99.97 ${\pm~ 0.029}$ & 99.98 ${\pm~ 0.018}$ & 99.97 ${\pm~ 0.002}$ & 99.97 ${\pm~ 0.026}$ & 99.86 ${\pm~ 0.084}$\\
& 2 & 99.80 & 99.85 & 99.75 ${\pm~ 0.127}$ & 99.80 ${\pm~ 0.126}$ & 99.71 ${\pm~ 0.122}$ & 99.74 ${\pm~ 0.052}$ & 99.82 ${\pm~ 0.027}$ \\
& 3 & 99.67 & 99.94 & 99.63 ${\pm~ 0.172}$ & 99.61 ${\pm~ 0.055}$ & 99.41 ${\pm~ 0.084}$ & 99.22 ${\pm~ 0.082}$ & 99.56 ${\pm~ 0.092}$ \\
(\%) & 4 & 99.49 & 100.0 & 99.05 ${\pm~ 0.470}$ & 99.15 ${\pm~ 0.032}$ & 98.61 ${\pm~ 0.312}$ & 97.84 ${\pm~ 0.180}$ & 98.80 ${\pm~ 0.292}$\\
& 5 & 99.10 & 99.86 & 99.00 ${\pm~ 0.100}$ & 99.06 ${\pm~ 0.171}$ & 97.31 ${\pm~ 0.575}$ & 96.77 ${\pm~ 0.337}$ & 98.63 ${\pm~ 0.430}$\\
\midrule
\multirow{4}{*}{$\boldsymbol{\alpha_{all,t}}$} & 1 & 100.0 & 100.0 &  99.97 ${\pm~ 0.029}$ & 99.98 ${\pm~ 0.018}$ & 99.97 ${\pm~ 0.002}$ & 99.97 ${\pm~ 0.026}$ & 99.86 ${\pm~ 0.084}$ \\
& 2 & 99.81 & 49.92 & 98.54 ${\pm~ 1.638}$ & 99.55 ${\pm~ 0.036}$  & 99.60 ${\pm~ 0.142}$ & 98.37 ${\pm~ 1.448}$ & 99.69 ${\pm~ 0.051}$\\
& 3 & 99.72 & 31.35 & 95.01 ${\pm~ 3.162}$ & 98.46 ${\pm~ 0.903}$  & 98.93 ${\pm~ 0.291}$ & 96.14 ${\pm~ 1.836}$ & 99.20 ${\pm~ 0.057}$ \\
(\%) & 4 & 99.50 & 24.82 & 81.50 ${\pm~ 9.369}$ & 97.06 ${\pm~ 1.069}$ & 98.22 ${\pm~ 0.560}$ & 91.25 ${\pm~ 0.992}$ & 98.13 ${\pm~ 0.281}$ \\
& 5 & 99.29 & 20.16 & 64.34 ${\pm~ 4.903}$ & 93.24 ${\pm~ 3.742}$ & 96.52 ${\pm~ 0.658}$ & 83.61 ${\pm~ 0.927}$ & 96.84 ${\pm~ 0.346}$\\
\midrule
\multirow{4}{*}{$\boldsymbol{\gamma_{base,t}}$} & 1 & 63.18 & 62.08 & 64.34 ${\pm~ 2.054}$ & 62.53 ${\pm~ 1.166}$ & 90.52 ${\pm~ 0.263}$ & 100.0 ${\pm~ 1.572}$ & 99.77 ${\pm~ 2.768}$ \\
& 2 & 62.85 & 126.8 & 74.41 ${\pm~ 10.89}$ & 65.68 ${\pm~ 1.166}$ & 91.27 ${\pm~ 0.789}$ & 100.4 ${\pm~ 1.964}$ & 101.2 ${\pm~ 3.601}$ \\
& 3 & 63.36 & 160.4 & 81.89 ${\pm~ 10.09}$ & 69.29 ${\pm~ 1.541}$ & 91.92 ${\pm~ 0.991}$ & 100.3 ${\pm~ 4.562}$ & 101.1 ${\pm~ 4.014}$ \\
(nats) & 4 & 64.25 & 126.9 & 90.62 ${\pm~ 10.08}$ & 71.69 ${\pm~ 1.379}$ & 91.75 ${\pm~ 1.136}$ & 102.7 ${\pm~ 7.134}$ & 101.0 ${\pm~ 4.573}$ \\
& 5 & 64.99 & 123.2 & 101.6 ${\pm~ 8.347}$ &  77.16 ${\pm~ 1.104}$  & 92.05 ${\pm~ 1.212}$ & 102.4 ${\pm~ 6.195}$ & 100.5 ${\pm~ 4.942}$ \\

\bottomrule
\end{tabular}}
\end{adjustwidth}
\end{table}

\begin{table}[H]\ContinuedFloat
\caption{{\em Cont.}}
\label{tab:incremental_MNIST_all}
\setlength{\tabcolsep}{3.2mm}
\begin{adjustwidth}{-\extralength}{0cm}
{\small\begin{tabular}{lllllllll}\toprule

\textbf{MNIST} & \textbf{\emph{t}} & \textbf{UB} & \textbf{FT} & \textbf{SupVAE} & \textbf{OpenVAE} & \textbf{PixelVAE DGR} & \textbf{SupPixelVAE} & \textbf{OpenPixelVAE} \\
\midrule

\multirow{4}{*}{$\boldsymbol{\gamma_{new,t}}$} & 1 & 63.18 & 62.08 & 64.34 ${\pm~ 2.054}$ & 62.53 ${\pm~ 1.166}$ & 90.52 ${\pm~ 0.263}$ & 100.0 ${\pm~ 1.572}$ & 99.77 ${\pm~ 2.768}$\\
& 2 & 88.75 & 87.93 & 89.91 ${\pm~ 0.107}$ & 89.64 ${\pm~ 3.709}$ & 115.8 ${\pm~ 0.805}$ & 125.7 ${\pm~ 2.413}$ & 124.6 ${\pm~ 3.822}$ \\
& 3 & 82.53 & 87.22 & 87.65 ${\pm~ 0.530}$ & 85.37 ${\pm~ 1.725}$ & 107.7 ${\pm~ 0.600}$ & 118.3 ${\pm~ 3.523}$ & 116.5 ${\pm~ 2.219}$ \\
(nats) & 4 & 72.68 & 74.61 & 79.49 ${\pm~ 0.489}$ & 74.75 ${\pm~ 0.777}$ & 100.9 ${\pm~ 0.659}$ & 107.1 ${\pm~ 5.316}$ & 102.3 ${\pm~ 1.844}$\\
& 5 & 85.88 & 92.00 & 93.55 ${\pm~ 0.391}$ &  89.68 ${\pm~ 0.618}$ & 113.4 ${\pm~ 0.820}$ & 118.2 ${\pm~ 1.572}$ & 113.3 ${\pm~ 0.755}$\\
\midrule
\multirow{4}{*}{$\boldsymbol{\gamma_{all,t}}$} & 1 & 63.18 & 62.08 & 64.34 ${\pm~ 2.054}$ & 62.53 ${\pm~ 1.166}$ & 90.52 ${\pm~ 0.263}$ & 100.0 ${\pm~ 1.572}$ & 99.77 ${\pm~ 2.768}$\\
& 2 & 75.97 & 107.3 & 82.02 ${\pm~ 5.488}$ & 76.62 ${\pm~ 1.695}$ & 102.9 ${\pm~ 0.408}$ & 111.9 ${\pm~ 2.627}$ & 112.7 ${\pm~ 3.300}$ \\
& 3 & 79.58 & 172.3 & 89.88 ${\pm~ 3.172}$ & 82.95 ${\pm~ 1.878}$ & 104.8 ${\pm~ 1.114}$ & 114.9 ${\pm~ 4.590}$ & 114.6 ${\pm~ 4.788}$ \\
(nats) & 4 & 79.72 & 203.1 & 95.83 ${\pm~ 2.747}$ & 85.30 ${\pm~ 1.524}$ & 103.9 ${\pm~ 0.759}$ & 114.3 ${\pm~ 3.963}$ & 112.1 ${\pm~ 2.150}$ \\
& 5 & 81.97 & 163.7 & 107.6 ${\pm~ 1.724}$ & 92.92 ${\pm~ 2.283}$ & 106.1 ${\pm~ 0.868}$ & 118.7 ${\pm~ 5.320}$ & 111.9 ${\pm~ 2.663}$ \\
\bottomrule
\end{tabular}}
\end{adjustwidth}
\end{table}
\unskip
\begin{table}[H]
\caption{The results for class incremental continual-learning approaches averaged over five runs, baselines and the reference isolated learning scenario for FashionMNIST at the end of every task increment. This is an extension of Table \ref{tab:incremental_results} in the main body. Here, in addition to the accuracy $\alpha_{t}$, $\gamma_{t}$ also indicates the respective NLL at the end of every task increment $t$. }
\label{tab:incremental_FashionMNIST_all}
\setlength{\tabcolsep}{3.2mm}
\begin{adjustwidth}{-\extralength}{0cm}
{\small\begin{tabular}{lllllllll}\toprule

\textbf{Fashion} & \textbf{\emph{t}} & \textbf{UB} & \textbf{FT} & \textbf{SupVAE} & \textbf{OpenVAE} & \textbf{PixelVAE DGR} & \textbf{SupPixelVAE} & \textbf{OpenPixelVAE}\\
\midrule

\multirow{4}{*}{$\boldsymbol{\alpha_{base,t}}$} & 1 & 99.65 & 99.60 & 99.55${\pm~ 0.035}$ & 99.59 ${\pm~ 0.082}$ & 99.57 ${\pm~ 0.091}$ & 99.58 ${\pm~ 0.076}$ & 99.54 ${\pm~ 0.079}$\\
& 2 & 96.70 & 00.00 & 92.02 ${\pm~ 1.175}$ & 92.36 ${\pm~ 2.092}$ & 82.40 ${\pm~ 6.688}$ & 90.06 ${\pm~ 1.782}$ & 88.60 ${\pm~ 1.998}$ \\
& 3 & 95.95 & 00.00 & 79.26 ${\pm~ 4.170}$ & 83.90 ${\pm~ 2.310}$ & 78.55 ${\pm~ 3.964}$ & 83.70 ${\pm~ 3.571}$ & 87.66 ${\pm~ 0.375}$ \\
(\%) & 4 & 91.35 & 00.00 & 50.16 ${\pm~ 6.658}$ & 64.70 ${\pm~ 2.580}$ & 54.69 ${\pm~ 3.853}$ & 50.23 ${\pm~ 7.004}$ & 68.31 ${\pm~ 3.308}$ \\
& 5 & 92.20 & 00.00 & 39.51 ${\pm~ 7.173}$ & 60.63 ${\pm~ 12.16}$ & 60.04 ${\pm~ 5.151}$ & 47.83 ${\pm~ 13.41}$ & 74.45 ${\pm~ 2.889}$ \\
\midrule
\multirow{4}{*}{$\boldsymbol{\alpha_{new,t}}$} & 1 & 99.65 & 99.60 & 99.55 ${\pm~ 0.035}$ & 99.59 ${\pm~ 0.082}$ & 99.57 ${\pm~ 0.091}$ & 99.58 ${\pm~ 0.076}$ & 99.54 ${\pm~ 0.079}$\\
& 2 & 95.55 & 97.95 & 90.98 ${\pm~ 0.626}$ & 92.64 ${\pm~ 2.302}$ & 97.73 ${\pm~ 1.113}$ & 96.47 ${\pm~ 0.596}$ & 97.31 ${\pm~ 0.475}$ \\
& 3 & 93.35 & 99.95 & 90.26 ${\pm~ 1.435}$ & 83.40 ${\pm~ 3.089}$ & 99.09 ${\pm~ 0.367}$ & 97.33 ${\pm~ 0.725}$ & 96.88 ${\pm~ 1.156}$\\
(\%) & 4 & 84.75 & 99.90 & 85.65 ${\pm~ 2.127}$ & 84.18 ${\pm~ 2.715}$ & 97.55 ${\pm~ 0.588}$ & 96.12 ${\pm~ 0.675}$ & 95.47 ${\pm~ 1.332}$\\
& 5 & 97.50 & 99.80 & 96.92 ${\pm~ 0.774}$ & 96.51 ${\pm~ 0.707}$ & 98.85 ${\pm~ 0.141}$ & 97.91 ${\pm~ 0.596}$ & 98.63 ${\pm~ 0.176}$ \\
%
%
\midrule
\multirow{4}{*}{$\boldsymbol{\alpha_{all,t}}$} & 1 & 99.65 & 99.60 & 99.55 ${\pm~ 0.035}$ & 99.59 ${\pm~ 0.082}$  & 99.57 ${\pm~ 0.091}$ & 99.58 ${\pm~ 0.076}$ & 99.54 ${\pm~ 0.079}$ \\
& 2 & 95.75 & 48.97 & 91.83 ${\pm~ 0.730}$ & 92.31 ${\pm~ 1.163}$ & 86.22 ${\pm~ 3.704}$ & 92.93 ${\pm~ 0.160}$ & 92.17 ${\pm~ 1.425}$\\
& 3 & 93.02 & 33.33 & 83.35 ${\pm~ 1.597}$ & 86.93 ${\pm~ 0.870}$ & 76.77 ${\pm~ 4.378}$ & 84.07 ${\pm~ 1.069}$ & 87.30 ${\pm~ 0.322}$ \\
(\%) & 4 & 87.51 & 25.00 & 64.66 ${\pm~ 3.204}$ & 76.05 ${\pm~ 1.391}$ & 62.93 ${\pm~ 3.738}$ & 64.42 ${\pm~ 1.837}$ & 76.36 ${\pm~ 1.267}$  \\
& 5 & 89.24 & 19.97 & 58.82 ${\pm~ 2.521}$ & 69.88 ${\pm~ 1.712}$ & 72.41 ${\pm~ 2.941}$ & 63.05 ${\pm~ 1.826}$ & 80.85 ${\pm~ 0.721}$ \\

\midrule

\multirow{4}{*}{$\boldsymbol{\gamma_{base,t}}$} & 1 & 209.7 & 209.8 & 208.9 ${\pm~ 1.213}$ & 209.7 ${\pm~ 3.655}$ & 267.8 ${\pm~ 1.246}$ & 230.8 ${\pm~ 3.024}$ & 232.0 ${\pm~ 2.159}$\\
& 2 & 207.4 & 240.7 & 212.7 ${\pm~ 0.579}$ & 212.1 ${\pm~ 0.937}$ & 273.6 ${\pm~ 0.631}$ & 232.5 ${\pm~ 1.582}$ & 231.8 ${\pm~ 0.416}$ \\
& 3 & 207.6 & 258.7 & 219.5 ${\pm~ 1.376}$ & 216.9 ${\pm~ 1.208}$ & 274.0 ${\pm~ 0.552}$ & 235.6 ${\pm~ 2.784}$ & 231.6 ${\pm~ 0.832}$ \\
(nats) & 4 & 207.7 & 243.6 & 223.8 ${\pm~ 0.837}$ & 217.1 ${\pm~ 0.979}$ & 273.7 ${\pm~ 0.504}$ & 236.4 ${\pm~ 3.157}$ & 231.4 ${\pm~ 2.550}$ \\
& 5 & 208.4 & 306.5 & 232.8 ${\pm~ 5.048}$ & 222.8 ${\pm~ 1.632}$ & 274.1 ${\pm~ 0.349}$ & 241.1 ${\pm~ 1.747}$ & 234.1 ${\pm~ 1.498}$ \\
\midrule
\multirow{4}{*}{$\boldsymbol{\gamma_{new,t}}$} & 1 & 209.7 & 209.8 & 208.9 ${\pm~ 1.213}$ & 209.7 ${\pm~ 3.655}$ & 267.8 ${\pm~ 1.246}$ & 230.8 ${\pm~ 3.024}$ & 232.0 ${\pm~ 2.159}$ \\
& 2 & 241.1 & 240.2 & 241.8 ${\pm~ 0.502}$ & 241.9 ${\pm~ 0.960}$ & 313.4 ${\pm~ 1.006}$ & 275.8 ${\pm~ 1.888}$ & 275.3 ${\pm~ 1.473}$ \\
& 3 & 213.6 & 211.8 & 215.4 ${\pm~ 0.501}$ & 213.0 ${\pm~ 0.635}$ & 269.1 ${\pm~ 0.616}$ & 268.3 ${\pm~ 3.852}$ & 262.9 ${\pm~ 1.893}$ \\
(nats) & 4 & 220.5 & 219.7 & 223.6 ${\pm~ 0.381}$ & 220.9 ${\pm~ 0.522}$ & 282.4 ${\pm~ 0.321}$ & 259.1 ${\pm~ 1.305}$ & 259.6 ${\pm~ 2.050}$  \\
& 5 & 246.2 & 242.0 & 248.8 ${\pm~ 0.398}$ & 244.0 ${\pm~ 0.646}$ & 305.8 ${\pm~ 0.286}$ & 283.2 ${\pm~ 2.150}$ & 283.5 ${\pm~ 2.458}$  \\
\midrule
\multirow{4}{*}{$\boldsymbol{\gamma_{all,t}}$} & 1 & 209.7 & 209.8 & 208.9 ${\pm~ 1.213}$ & 209.7 ${\pm~ 3.655}$ & 267.8 ${\pm~ 1.246}$ & 230.8 ${\pm~ 3.024}$ & 232.0 ${\pm~ 2.159}$ \\
& 2 & 224.2 & 240.4 & 226.6 ${\pm~ 2.31}$ & 226.9 ${\pm~ 0.918}$ & 293.8 ${\pm~ 0.349}$ & 254.3 ${\pm~ 1.513}$ & 255.8 ${\pm~ 0.436}$ \\
& 3 & 220.7 & 246.1 & 227.2 ${\pm~ 0.606}$ & 224.9 ${\pm~ 0.642}$ & 285.7 ${\pm~ 0.510}$ & 261.5 ${\pm~ 2.970}$ & 259.1 ${\pm~ 0.929}$ \\
(nats) & 4 & 220.4 & 238.7 & 230.4 ${\pm~ 0.524}$ & 226.1 ${\pm~ 0.560}$ & 284.9 ${\pm~ 0.703}$ & 263.2 ${\pm~ 2.259}$ & 259.5 ${\pm~ 3.218}$  \\
& 5 & 226.2 & 275.1 & 242.2 ${\pm~ 0.754}$ & 234.6 ${\pm~ 0.823}$ & 289.5 ${\pm~ 0.396}$ & 271.7 ${\pm~ 2.117}$ & 267.2 ${\pm~ 0.586}$\\
\bottomrule
\end{tabular}}
\end{adjustwidth}
\end{table}
\unskip
\begin{table}[H]
\caption{The results for class incremental continual-learning approaches averaged over five runs, baselines and the reference isolated learning scenario for AudioMNIST at the end of every task increment. This is an extension of Table \ref{tab:incremental_results} in the main body. Here, in addition to the accuracy $\alpha_{t}$, $\gamma_{t}$ also indicates the respective NLL at the end of every task increment $t$. }
\label{tab:incremental_AudioMNIST_all}
\setlength{\tabcolsep}{3.3mm}
\begin{adjustwidth}{-\extralength}{0cm}
{\small\begin{tabular}{lllllllll}\toprule

\textbf{Audio} & \textbf{\emph{t}} & \textbf{UB} & \textbf{LB} & \textbf{SupVAE} & \textbf{OpenVAE} & \textbf{PixelVAE DGR} & \textbf{SupPixelVAE} & \textbf{OpenPixelVAE}\\
\midrule

\multirow{4}{*}{$\boldsymbol{\alpha_{base,t}}$} & 1 & 99.99 & 100.0 & 99.21${\pm~ 0.568}$ & 99.95 ${\pm~ 0.035}$ & 100.0 ${\pm~ 0.000}$ & 99.71 ${\pm~ 0.218}$ & 99.27 ${\pm~ 0.410}$ \\
& 2 & 99.92 & 00.00 & 98.98 ${\pm~ 0.766}$ & 98.61 ${\pm~ 0.490}$ & 99.52 ${\pm~ 0.273}$ & 97.86 ${\pm~ 0.799}$ & 97.88 ${\pm~ 2.478}$ \\
& 3 & 100.0 & 00.00 & 92.44 ${\pm~ 1.306}$ & 95.12 ${\pm~ 2.248}$ & 93.15 ${\pm~ 3.062}$ & 81.38 ${\pm~ 5.433}$ & 95.82 ${\pm~ 3.602}$ \\
(\%) & 4 & 99.92 & 00.00 & 76.43 ${\pm~ 4.715}$ & 86.37 ${\pm~ 5.63}$ & 81.55 ${\pm~ 8.468}$ & 50.58 ${\pm~ 14.60}$ & 91.56 ${\pm~ 5.640}$  \\
& 5 & 98.42 & 00.00 & 59.36 ${\pm~ 7.147}$ & 79.73 ${\pm~ 4.070}$ & 64.60 ${\pm~ 8.739}$ & 29.94 ${\pm~ 18.47}$ & 75.25 ${\pm~ 10.18}$ \\
\midrule
\multirow{4}{*}{$\boldsymbol{\alpha_{new,t}}$} & 1 & 99.99 & 100.0 & 99.21 ${\pm~ 0.568}$ & 99.95 ${\pm~ 0.035}$ & 100.0 ${\pm~ 0.000}$ & 99.71 ${\pm~ 0.218}$ & 99.27 ${\pm~ 0.410}$ \\
& 2 & 99.75 & 100.0 & 91.82 ${\pm~ 4.577}$ & 89.23 ${\pm~ 7.384}$ & 99.71 ${\pm~ 0.043}$ & 99.78 ${\pm~ 0.128}$ & 99.81 ${\pm~ 0.189}$ \\
& 3 & 98.92 & 99.58 & 95.20 ${\pm~ 1.495}$ & 94.43 ${\pm~ 3.030}$ & 98.23 ${\pm~ 1.092}$ & 98.41 ${\pm~ 0.507}$ & 99.30 ${\pm~ 0.550}$ \\
(\%) & 4 & 97.33 & 98.67 & 53.02 ${\pm~ 6.132}$ & 72.22 ${\pm~ 8.493}$ & 95.31 ${\pm~ 0.868}$ & 94.30 ${\pm~ 0.914}$ & 97.87 ${\pm~ 0.293}$ \\
& 5 & 98.67 & 100.0 & 84.93 ${\pm~ 6.297}$ & 89.52 ${\pm~ 6.586}$ & 98.18 ${\pm~ 0.885}$ & 97.00 ${\pm~ 0.520}$ & 99.43 ${\pm~ 0.495}$  \\

\midrule
\multirow{4}{*}{$\boldsymbol{\alpha_{all,t}}$} & 1 & 99.99 & 100.0 & 99.21 ${\pm~ 0.568}$ & 99.95 ${\pm~ 0.035}$ & 100.0 ${\pm~ 0.000}$ & 99.71 ${\pm~ 0.218}$ & 99.27 ${\pm~ 0.410}$ \\
& 2 & 99.83 & 50.00 & 93.84 ${\pm~ 2.558}$ & 93.93 ${\pm~ 3.756}$ & 99.50 ${\pm~ 0.157}$ & 98.64 ${\pm~ 0.875}$ & 99.67 ${\pm~ 0.033}$ \\
& 3 & 99.56 & 33.19 & 94.26 ${\pm~ 1.669}$ & 95.70 ${\pm~ 1.524}$ & 95.37 ${\pm~ 1.750}$ & 90.10 ${\pm~ 1.431}$ & 97.77 ${\pm~ 1.017}$  \\
(\%) & 4 & 98.60 & 24.58 & 77.90 ${\pm~ 4.210}$ & 85.59 ${\pm~ 3.930}$ & 86.97 ${\pm~ 2.797}$ & 75.55 ${\pm~ 3.891}$ & 95.41 ${\pm~ 1.345}$\\
& 5 & 97.87 & 20.02 & 81.49 ${\pm~ 1.944}$ & 87.72 ${\pm~ 1.594}$ & 75.50 ${\pm~ 3.032}$ & 63.44 ${\pm~ 5.252}$ & 90.23 ${\pm~ 1.139}$ \\
\midrule
\multirow{4}{*}{$\boldsymbol{\gamma_{base,t}}$} & 1 & 433.7 & 423.2 & 435.2 ${\pm~ 15.69}$ & 424.2 ${\pm~ 2.511}$ & 434.2 ${\pm~ 1.068}$ & 432.6 ${\pm~ 0.321}$ & 433.8 ${\pm~ 0.370}$ \\
& 2 & 422.5 & 439.4 & 423.9 ${\pm~ 0.517}$ & 425.2 ${\pm~ 1.402}$ & 434.4 ${\pm~ 1.082}$ & 432.5 ${\pm~ 0.551}$ & 433.5 ${\pm~ 1.464}$\\
& 3 & 420.7 & 429.2 & 422.7 ${\pm~ 0.690}$ & 423.8 ${\pm~ 1.148}$ & 434.6 ${\pm~ 0.785}$ & 432.9 ${\pm~ 0.723}$ & 433.1 ${\pm~ 1.269}$ \\
(nats) & 4 &419.9 & 428.5 & 422.8 ${\pm~ 0.367}$ & 423.5 ${\pm~ 0.937}$ & 434.2 ${\pm~ 1.209}$ & 433.0 ${\pm~ 0.781}$ & 433.0 ${\pm~ 1.283}$\\
& 5 & 418.4 & 432.9 & 422.7 ${\pm~ 0.182}$ & 423.5 ${\pm~ 0.586}$ & 435.1 ${\pm~ 1.915}$ & 431.4 ${\pm~ 0.666}$ & 432.3 ${\pm~ 0.189}$ \\

\midrule

\multirow{4}{*}{$\boldsymbol{\gamma_{new,t}}$} & 1 & 433.7 & 423.2 & 435.2 ${\pm~ 15.69}$ & 424.2 ${\pm~ 2.511}$  & 434.2 ${\pm~ 1.068}$ & 432.6 ${\pm~ 0.321}$ & 433.8 ${\pm~ 0.370}$\\
& 2 & 381.2 & 384.1 & 382.5 ${\pm~ 1.355}$ & 385.3 ${\pm~ 12.56}$ & 390.4 ${\pm~ 0.694}$ & 389.4 ${\pm~ 0.208}$ & 389.4 ${\pm~ 1.304}$ \\
& 3 & 435.9 & 436.7 & 436.3 ${\pm~ 0.639}$ & 436.9 ${\pm~ 0.688}$ & 444.7 ${\pm~ 0.545}$ & 442.7 ${\pm~ 0.513}$ & 442.4 ${\pm~ 0.275}$ \\
(nats) & 4 & 485.9 & 487.1 & 486.7 ${\pm~ 0.385}$ & 486.5 ${\pm~ 0.701}$ & 497.4 ${\pm~ 0.740}$ & 494.4 ${\pm~ 0.700}$ & 494.8 ${\pm~ 0.386}$\\
& 5 & 421.3 & 425.2 & 423.9 ${\pm~ 0.681}$ & 422.9 ${\pm~ 0.537}$ & 431.9 ${\pm~ 1.032}$ & 428.0 ${\pm~ 0.851}$ & 429.7 ${\pm~ 1.223}$ \\

\midrule
\multirow{4}{*}{$\boldsymbol{\gamma_{all,t}}$} & 1 & 433.7 & 423.2 & 435.2 ${\pm~ 15.69}$ & 424.2 ${\pm~ 2.511}$ & 435.2 ${\pm~ 15.69}$ & 432.6 ${\pm~ 0.321}$ & 433.8 ${\pm~ 0.370}$ \\
& 2 & 401.9 & 411.8 & 403.2 ${\pm~ 0.831}$ & 403.5 ${\pm~ 1.274}$ & 412.4 ${\pm~ 0.871}$ & 410.9 ${\pm~ 0.351}$ & 411.5 ${\pm~ 1.406}$  \\
& 3 & 412.1 & 418.9 & 413.6 ${\pm~ 0.410}$ & 413.8 ${\pm~ 0.573}$ & 423.3 ${\pm~ 0.618}$ & 421.0 ${\pm~ 1.026}$ & 421.9 ${\pm~ 0.661}$  \\
(nats) & 4 & 430.3 & 438.4 &  432.4 ${\pm~ 0.436}$ & 432.6 ${\pm~ 0.862}$ & 441.6 ${\pm~ 0.420}$ & 439.8 ${\pm~ 0.833}$ & 439.8 ${\pm~ 0.718}$  \\
& 5 & 427.2 & 440.4 & 431.4 ${\pm~ 0.255}$ & 430.9 ${\pm~ 0.541}$  & 440.3 ${\pm~ 1.297}$ & 436.9 ${\pm~ 0.751}$ & 437.7 ${\pm~ 0.432}$ \\
\bottomrule
\end{tabular}}
\end{adjustwidth}
\end{table}

\reftitle{References}



\begin{adjustwidth}{-\extralength}{0cm}

\end{adjustwidth}

%


%
%
%
\end{document}